\documentclass{article}

\usepackage[preprint]{neurips_2025}

\usepackage{amsmath,amssymb,amsfonts}
\usepackage{algorithm,algorithmic}

\usepackage{enumitem}
\usepackage[pdftex]{graphicx}
\usepackage{pifont}
\newcommand{\cmark}{\ding{51}} 
\newcommand{\xmark}{\ding{55}} 
\usepackage{multirow}
\usepackage[utf8]{inputenc} 
\usepackage[T1]{fontenc}    
\usepackage{hyperref}       
\usepackage{url}            
\usepackage{booktabs}       
\usepackage{amsfonts}       
\usepackage{nicefrac}       
\usepackage{microtype}      
\usepackage{xcolor}         
\usepackage{graphicx}

\title{LLM-to-Phy3D: Physically Conform Online 3D Object Generation with LLMs}

\author{%
  Melvin Wong\(^{1*}\) \quad Yueming Lyu\(^{2}\) \quad Thiago Rios\(^{3}\) \quad Stefan Menzel\(^{3}\) \quad Yew-Soon Ong\(^{1,2}\)\thanks{corresponding authors} \\ [8pt]
  \(^{1}\)College of Computing \& Data Science, Nanyang Technological University, Singapore \\ [3pt]
  \(^{2}\)Centre for Frontier AI Research, Agency for Science, Technology and Research, Singapore \\ [3pt]
  \(^{3}\)Honda Research Institute Europe (HRI-EU), Offenbach am Main, Germany \\ [8pt]
  \texttt{\{wong1357, asysong\}@ntu.edu.sg} \\
  \texttt{Lyu\_Yueming@cfar.a-star.edu.sg} \\
  \texttt{\{thiago.rios, stefan.menzel\}@honda-ri.de}
}

\begin{document}
\maketitle

\begin{abstract}

    The emergence of generative artificial intelligence (GenAI) and large language models (LLMs) has revolutionized the landscape of digital content creation in different modalities. However, its potential use in Physical AI for engineering design, where the production of physically viable artifacts is paramount, remains vastly underexplored. The absence of physical knowledge in existing LLM-to-3D models often results in outputs detached from real-world physical constraints. To address this gap, we introduce LLM-to-Phy3D, a physically conform online 3D object generation that enables existing LLM-to-3D models to produce physically conforming 3D objects on the fly. LLM-to-Phy3D introduces a novel online black-box refinement loop that empowers large language models (LLMs) through synergistic visual and physics-based evaluations. By delivering directional feedback in an iterative refinement process, LLM-to-Phy3D actively drives the discovery of prompts that yield 3D artifacts with enhanced physical performance and greater geometric novelty relative to reference objects, marking a substantial contribution to AI-driven generative design. Systematic evaluations of LLM-to-Phy3D, supported by ablation studies in vehicle design optimization, reveal various LLM improvements gained by 4.5\% to 106.7\% in producing physically conform target domain 3D designs over conventional LLM-to-3D models. The encouraging results suggest the potential general use of LLM-to-Phy3D in Physical AI for scientific and engineering applications.

\end{abstract}

\begin{figure}
    \centering
    \includegraphics[width=1.00\textwidth]{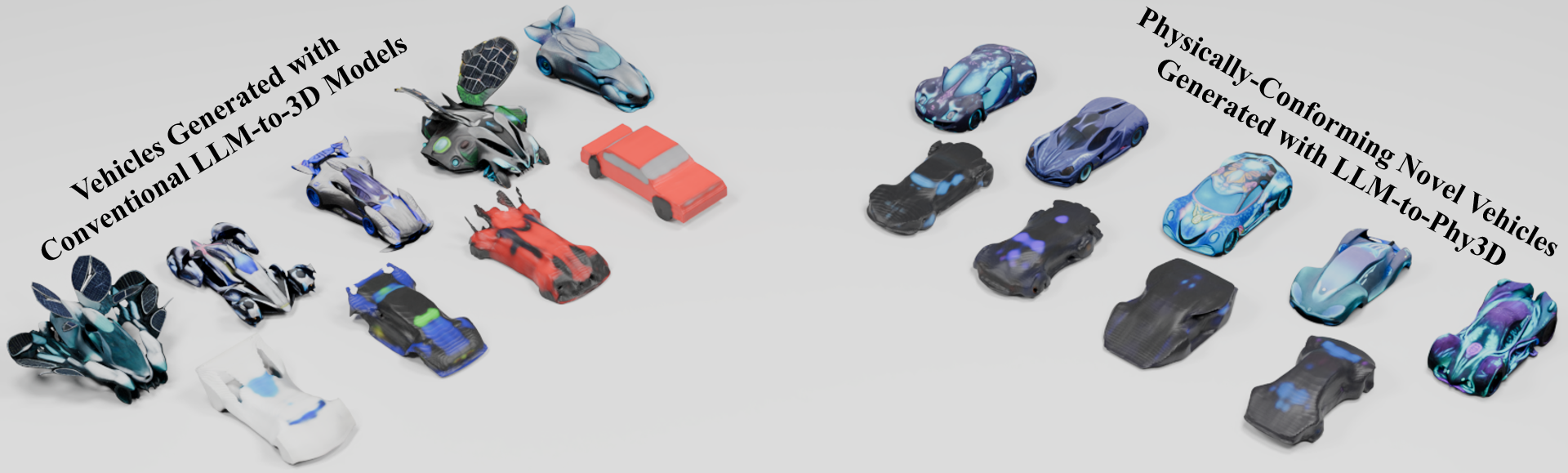} \\
    \caption{Most representative vehicles generated with various conventional LLM-to-3D models using LLM-generated prompts (left) and physically conforming novel vehicles generated with LLM-to-Phy3D (right).}
    \label{fig:showcase}
\end{figure}

\section{Introduction} \label{sec:intro}


Remarkable advancements in generative artificial intelligence (GenAI) and large language models (LLMs) have opened up new research opportunities, focusing primarily on the creation of high-fidelity digital artifacts that faithfully align with user preferences through tailored prompts. This has led to a surge in broad applications including text-to-\emph{X} generation that produces digital artifacts of different modalities (e.g., text, image, or 3D) conditioned on textual prompts (\cite{dong2022nlg,bei2025tti,lee2024text}).

Among those applications, text-to-3D object generation has attracted increasing attention due to its success and potential to democratize 3D content creation, accelerate design workflows, and leverage the capabilities of emerging multimodal foundation models (\cite{lee2024text}). One main problem of 3D object generation is automating the design process while enabling the synthesis of novel and diverse shapes (\cite{Li2024AdvancesI3}). Several works introduce text-to-3D models that can generate the parameters of implicit functions for rendering textured meshes and neural radiance fields (\cite{li2023diffusion,jun2023shap}). This enables the synthesis of diverse 3D shapes in a target domain from a single text prompt. \cite{xiang2024structured} proposed a unified structured latent representation that expands the generation of outputs to the 3D Gaussian format and improves the quality of generated artifacts. \cite{wang2024llama} attempted to unify the spatial textual knowledge in LLMs with 3D information to improve mesh understanding, enabling the generation of simple polygonal meshes in textual format.

\begin{figure}
  \centering
  \includegraphics[width=1.00\linewidth]{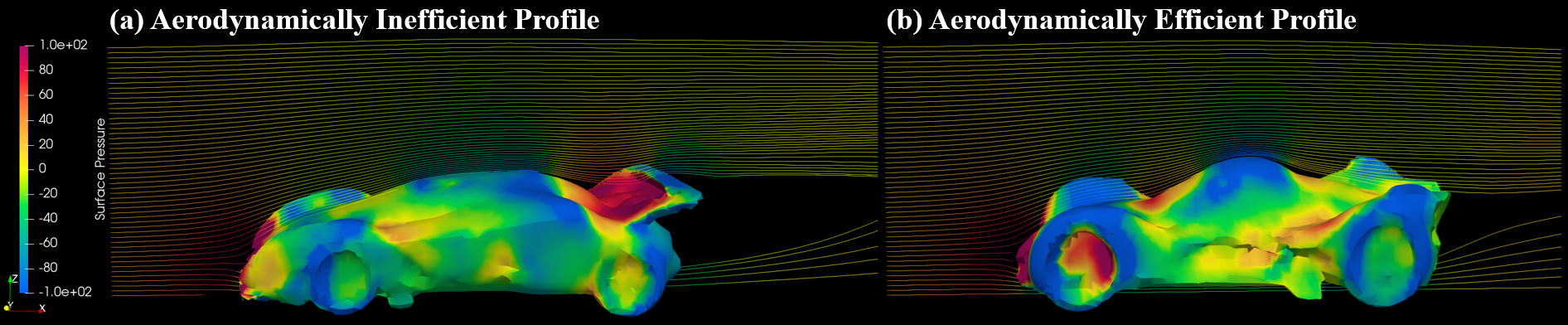}
  \caption{Aerodynamic Visualization of two vehicles generated (a) without considering physical constraints and (b) with physical constraints. Vehicle shape has a direct impact on the aerodynamic (drag) performance. Some aerodynamically efficient (low drag) characteristics include attached streamline flow from the front to the rear, and small pressure differences between the front and rear. In contrast, an aerodynamically inefficient (high drag) vehicle exhibits opposing characteristics, including separated streamline flow and pronounced pressure differences.}
  \label{fig:comparison}
\end{figure}

While these advancements led to the proliferation of digital media applications, there is a notable lack of progress in advancing 3D object generation in Physical AI for engineering design, where the production of physically viable artifacts is paramount (\cite{banerjee2024physics,liu2025generative}). One major challenge we observed is the generation of objects that satisfy physical constraints. 3D vehicles generated from conventional text-to-3D models without physical knowledge (Figure~\ref{fig:showcase} (Left Group) and Figure~\ref{fig:comparison} (a)) while aesthetically pleasing, have limited efficacy in both the real physical world and its digital twin as compared to ones generated with these constraints (Figure~\ref{fig:showcase} (Right Group) and Figure~\ref{fig:comparison} (b)). Notable work attempts to incorporate such constraints in GenAI models, focusing on enforcing the generation of a targeted object shape to comply with physical laws (\cite{xu2025fun3d}). However, due to confidentiality concerns and prohibitive data generation costs, the scarcity of large-scale, high-quality engineering data poses difficulties in scaling and generalizing such techniques for text-to-3D generation. In contrast, black-box techniques can overcome such difficulties, enabling existing GenAI models to generate novel artifacts that conform to targeted physical constraints. Pioneering research studies that investigate this approach in engineering design optimization, where preliminary results, particularly in enhancing aerodynamic performance while satisfying domain constraints (\cite{rios2023large,wong2024prompt}), encourage progress and interest. These efforts motivate for GenAI with language models to narrow the gap between digital innovation and functional applicability, driving progress in Physical AI for engineering design. However, these methods focus on improving conventional optimization strategies. To date, the feasibility of LLMs in producing 3D artifacts that conform to physics for engineering design remains vastly underexplored.


To this end, we introduce \textit{LLM-to-Phy3D}, a novel physically conform online 3D object generation with LLMs. Central to LLM-to-Phy3D is its unique online black-box iterative refinement framework that adaptively steers the LLM to find textual prompts that yield physically plausible 3D artifacts with high geometric novelty. The framework starts with 3D artifacts generated with randomly sampled LLM-generated prompts. Subsequently, LLM-to-Phy3D quantifies the physical performance, target domain alignment, and geometric novelty of 3D artifacts with visual and physics-based evaluations, which serve as surrogate indicators for a non-differentiable black-box objective. An iteration of LLM-to-Phy3D ends with a selection process that picks the best-performing LLM-generated prompts as exemplars for the LLM to learn the relationship between its prompts and the generated 3D artifacts that resulted in the quantified score, enabling the LLM to improve its performance in the next iteration. This selection pressure mechanism not only ensures constant space complexity but, more importantly, achieves the desired generation characteristics with directional feedback (\cite{shimabucoro2024llm,nie2023importance}) that steers the LLM to produce physically conforming target domain 3D artifacts.

Moreover, generation issues, such as hallucinations (\cite{jesson2024estimating}), produce 3D artifacts with shapes that are not within the target domain yet satisfy the physical constraints. Visual evaluations can not only effectively identify and penalize irrelevant shapes but also reward novel ones. While existing foundation vision models can assess such artifacts from 2D rendered views with a high degree of accuracy (\cite{li2023blip,tan2019efficientnet}), the projection of 3D artifacts onto 2D can cause significant geometric distortions in the surface topology that degrade the model performance. To address such issues, we propose conducting the physical light simulation under orthographic projection (\cite{parker2010optix}) to capture the precise surface topology of 3D artifacts and minimize geometric distortions. This single visual representation serves the dual purpose of object recognition and geometric novelty quantification with respect to reference objects.





To demonstrate the efficacy of LLM-to-Phy3D against conventional LLM-to-3D methods, we conduct experiments on a test scenario that mimics real-world engineering design optimization. Starting from initial LLM-generated prompts that yield 3D artifacts with significant variation in physical performance, we demonstrate that all LLMs in our experiments are able to improve the quality of its generated prompts, leading to significant improvements in producing physically conforming novel 3D artifacts with better realism than the ones generated with conventional text-to-3D models. To summarize, the contributions of this paper are as follows:

\begin{itemize}[leftmargin=2em]
    \vspace{-0.5em}
    \item We introduce \textit{LLM-to-Phy3D}, a novel physically conform online 3D object generation with LLMs, for actively steering the LLM to produce physically conforming novel 3D objects. LLM-to-Phy3D enables the LLM to find effective prompts by providing physical knowledge through directional feedback. This allows the LLM to learn the association between its generated prompts and the quality of 3D artifacts generated with its these prompts.

    \item LLM-to-Phy3D captures the precise surface topology and visual characteristics of 3D artifacts through physical light simulation with orthographic projection in a single visual representation for object recognition and geometric novelty quantification with respect to reference objects. The technique minimizes geometric distortion, ensuring the rendered surface topology conforms to scientific and engineering requirements.

    \item Visual and physics-based evaluations are introduced to quantify the quality of 3D-generated artifacts based on its physical performance and geometric novelty. These evaluations contribute to the search objective, enabling the selection of physically conforming ones as exemplars for the LLM to learn from in the next iteration.

    \item In contrast to existing LLM benchmark approaches, which tailor meta-prompts for different LLMs, we evaluate the performance of four state-of-the-art LLMs with the same standardized meta-prompt to ensure a fair comparison of search performance. Systematic evaluations of LLM-to-Phy3D, along with ablation studies in aerodynamic vehicle design optimization tasks, reveal various LLM improvements gained by 4.5\% to 106.7\% in producing physically conforming target domain 3D designs, indicating the broader potential of the framework in Physical AI for science and engineering applications.
\end{itemize}
\vspace{-1em}
\vspace{-0.5em}
\section{Preliminaries}
\vspace{-1em}
In this paper, we articulate the physical viability of 3D objects in the context of, but not limited to, the automotive aerodynamics design scenario. As such, we introduce in this section the key fundamental concepts in representing and quantifying the novelty and quality of 3D objects.



\textbf{Surface Topology Representation for Object Recognition and Geometric Novelty Quantification:} In many computer vision and computer graphics applications, the semantic meaning and visual features of a 3D object are defined based on the object's surface topology (\cite{rothwell1996representing,montagnat2001review,pulli2000surface}). This topology information can be captured with the illumination of the surface through a physical light simulation (\cite{parker2010optix}), which replicates physical light rays reflecting and refracting from the surface and projects onto a camera plane. Assuming camera orthographic projection, given a set of all intersection points \(\textbf{Z}_{c}\) in a fixed camera viewpoint \(c\) of a 3D-object \(\textbf{x}\), the surface topology of the rendered \(R\) multi-view is defined as:
\begin{equation} \label{eq:mv}
    g(\textbf{x}) = \{\textbf{X}=\{I(\textbf{z}) \, | \, \forall \textbf{z} \in \textbf{Z}_{c}\} \, | \, \forall c \in \{1,...,R\}\},
\end{equation}
where \(\textbf{X}\) is a rendered view image of \(\textbf{x}\) consisting of the illumination (or shading) from multiple light rays intersecting at different surface point \(\textbf{z}\) by \(I(\cdot)\). In this paper, we compare the multi-view of generated and reference 3D objects to minimize inaccuracies in determining the region of geometric novelty due to surface topology distortion caused by indirect illumination. Details of this illumination process (\cite{dutre2018advanced}) are discussed in the supplementary background Section~\ref{sec:rendering}. 

\textbf{Aerodynamics Computation for Physical Performance Quantification:} For automotive aerodynamics, the drag and lift characteristics of a vehicle are of prime interest. These characteristics can be quantified with a physics simulator (or surrogate) \(G\) where the vehicle is subject to fluid flow conditions, such as steady or unsteady air streams, replicating real-world aerodynamic environments. The fluid particles interacting with the surface of the vehicle create viscous forces and pressures in the flow direction that drag or lift the vehicle body. Assuming the particles are travelling along the \(x\)-axis, given the projected area of the vehicle (\(A_{\text{projected}}\)), the drag and lift scores are defined as:
\vspace{-0.5em}
\begin{equation} \label{eq:drag}
    \begin{aligned}
        & G_{\text{fluid\_forces}}(\textbf{x}, \mathcal{F}) \approx \frac{\mathcal{F}(\textbf{x})}{2\rho_{\infty}\|\textbf{u}_{\infty}\|_2^{2} \cdot A_{\text{projected}}}, \\
        & \bar{C}_{\text{lift}}(\textbf{x}) = G_{\text{fluid\_forces}}(\textbf{x}, \mathcal{F}_{z}),
        & \bar{C}_{\text{drag}}(\textbf{x}) = G_{\text{fluid\_forces}}(\textbf{x}, \mathcal{F}_{x}),
    \end{aligned}
\end{equation}
where \(\rho_{\infty}\) is the freestream ambient density of the fluid, and \(\textbf{u}_{\infty}\) is the freestream velocity vector. Here, \(\mathcal{F}(\cdot)\) computes the integral forces acting on the surface with \(\mathcal{F}_{x}(\cdot)\) considering the forces along the \(x\)-axis only while \(\mathcal{F}_{z}(\cdot)\) take into account forces along the \(z\)-axis only. Note that the drag and lift coefficients (\(\bar{C}_{\text{drag}}\) and \(\bar{C}_{\text{lift}}\)) are usually computed in a dimensionless and normalized form. Background information on the fluid dynamics computation (\cite{lomax2001fundamentals}) to compute the drag and lift is presented in the supplementary background Section~\ref{sec:drag}. 

\vspace{-1em}
\section{Method: LLM-to-Phy3D} \label{sec:methods}
\vspace{-0.5em}
We first present the LLM-to-Phy3D framework, followed by detailed discussions of the key components of the framework.
\vspace{-0.5em}
\subsection{Online Blackbox Iterative Refinement Framework}
\vspace{-0.5em}


Without loss of generality, let \(\mathcal{P}\) denote a space of textual prompts and \(\mathcal{X}\) denote a space of 3D objects. Given any 3D object \(\textbf{x} \in \mathcal{X}\), we assume a text-to-3D GenAI model (\(p_{\text{t3d}}\)) that generate new 3D objects conditioned on a textual prompt. Moreover, given any prompt \(\textbf{m} \in \mathcal{P}\), we also assume the presence of an LLM-based optimizer (\(p_{\text{llm}}\)) that can iteratively generate new prompts based on the domain of interest \(\mathcal{S}\) and a meta-prompt \(\mathbb{M}\), and can be adaptively guided using exemplars consisting of prompts paired with its performance in generating high-quality 3D objects. As such, given a set \(\textbf{Y}\) of reference 3D objects, the search goal is to find novel 3D artifacts that satisfy physical and domain constraints: \vspace{-0.5em}
\begin{equation} \label{eq:obj}
    \begin{aligned}
        & \min_{\boldsymbol{\textbf{m}} \in \mathcal{P}}: \mathop{\mathbb{E}}_{\textbf{x} \sim p_{\text{t3d}}(\textbf{x} \, | \, \boldsymbol{\textbf{m}})} \left[ f_{\text{physical}}(\textbf{x}) + f_{\text{domain}}(g(\textbf{x}), \mathcal{S}) - \beta F_{\text{novelty}}(\textbf{x}, \textbf{Y}) \right] \\
        & \text{s.t.} \,\,\,\, Q(\textbf{m}, \mathcal{S}) \le \epsilon, \\
        & \quad\quad \underline{c} \le f_{\text{physical}}(\textbf{x}) \le \bar{c},
    \end{aligned}
\end{equation}
where \(Q(\cdot, \cdot)\) is a function that filters infeasible LLM-generated prompts, due to generation failure scenarios (e.g. LLM hallucination), which are semantically distant to the domain of interest \(\mathcal{S}\) by \(\epsilon\). Moreover, \(\underline{c}>0.0\) and \(\bar{c} < 1.0\) define the boundary that any 3D generated artifact in the target domain can realistically achieve in the real world, effectively filtering infeasible ones such as the non-conforming or non-watertight with object's surfaces are not fully closed. The first term ranks generated artifacts with better physical performance, scoring higher than impractical ones. Among those that satisfy the physical constraint, generated artifacts less resembling \(\mathcal{S}\) are penalized in the second term, demoting its position for such artifacts. Finally, the third term rewards artifacts with high geometric novelty with respect to the target domain, promoting the positions. This term is weighted with a \(\beta\) parameter to balance the novelty score with the physical score. These three terms work in synergy to provide the LLM with effective feedback on the quality of generated artifacts, ensuring that the better-quality candidates serve as exemplars to the LLM in subsequent iterations.

Algorithm~\ref{alg:fit3d} describes the mechanics of the framework. Given a user domain \(\mathcal{S}\) and design specifications \(\mathcal{C}\), LLM-to-Phy3D initializes the LLM (\(p_{\text{llm}}\)) with the search task. Our proposed method then samples prompts from the LLM to produce the corresponding 3D artifacts from the text-to-3D generative model \(p_{\text{t3d}}\). The generated 3D artifacts are evaluated against the specifications to determine the scores. These scores are then combined with the LLM-generated prompts to serve as exemplars for the LLM to learn and re-attempt to create new prompts that can yield better-performing candidates. These candidates are evaluated and, together with the ones generated in the last iteration, undergo a selection process. This process selects the better-performing candidate from randomly paired candidates to form an exemplar set of \(N\) candidates for the next iteration (Line 7 in Algorithm~\ref{alg:fit3d}), ensuring constant space complexity throughout the search. Moreover, the exemplar set serves as directional feedback, allowing it to learn the association between the prompt and the quality of the 3D object and find more effective prompts that steer towards the optimal. LLM-to-Phy3D iterates until the terminating condition is satisfied. \vspace{-1em}
\begin{algorithm}
    \caption{LLM-to-Phy3D}\label{alg:fit3d}
    \begin{algorithmic}[1]
        \REQUIRE \(\mathbb{M}\) meta prompt, \(\mathcal{S}\) domain specifications, and \(\mathcal{C}\) design specifications, batch-size \(N\)
        \ENSURE Designs Set \(\mathcal{D}_{\text{step}}\)
        \STATE \(\text{step} \leftarrow  0\)
        \STATE \(\mathcal{D}_{\text{step}} \leftarrow \{(\boldsymbol{\textbf{m}}_{i}, \textbf{x}_{i}) \, | \, \boldsymbol{\textbf{m}}_{i} \sim p_{\text{llm}}(\boldsymbol{\textbf{m}}_{i} \, | \, \mathbb{M}, \mathcal{S}), \, \textbf{x}_{i} \sim p_{\text{t3d}}(\textbf{x}_{i} \, | \, \boldsymbol{\textbf{m}}_{i})\} \quad \forall i \in \{1,...,N\}\)
        \STATE Evaluate \(\mathcal{D}_{\text{step}}\) against \(\mathcal{C}\) and \(\mathcal{S}\) based on Equation (\ref{eq:domain}), (\ref{eq:physical}), (\ref{eq:topomin}), and (\ref{eq:obj})
        \WHILE {non-terminating condition}
            \STATE \(\mathcal{D}_{\text{candidates}} \leftarrow \{(\boldsymbol{\textbf{m}}_{i}, \textbf{x}_{i}) \, | \, \boldsymbol{\textbf{m}}_{i} \sim p_{\text{llm}}(\boldsymbol{\textbf{m}}_{i} \, | \, \mathbb{M}, \mathcal{S}, \mathcal{D}_{\text{step}}), \, \textbf{x}_{i} \sim p_{\text{t3d}}(\textbf{x}_{i} \, | \, \boldsymbol{\textbf{m}}_{i})\} \quad \forall i \in \{1,...,N\}\)
            \STATE Evaluate \(\mathcal{D}_{\text{candidates}}\) against \(\mathcal{C}\) and \(\mathcal{S}\) based on Equation (\ref{eq:domain}), (\ref{eq:physical}), (\ref{eq:topomin}), and (\ref{eq:obj})
            \STATE \(\mathcal{D}_{\text{step+1}} \leftarrow \text{select\_N\_candidates}(\mathcal{D}_{\text{candidates}} \cup \mathcal{D}_{\text{step}}) \)
            \STATE \(\text{step} \leftarrow \text{step} + 1\)
        \ENDWHILE
    \end{algorithmic}
\end{algorithm}
\subsection{Quality of 3D Generated Objects}
\vspace{-0.5em}
We quantify the quality of a 3D-generated object \(\textbf{x}\) with the following measures:

\textbf{(a) Physical Constraint with Physical Alignment Measure}: This measure assesses the performance of 3D-generated objects under various physics conditions, governed by the natural physical laws. Here, we provide an instantiation of this measure in the context of fluid dynamics (\cite{lomax2001fundamentals}). As such, the physical alignment measure is defined as follows: 
\begin{equation} \label{eq:physical}
    f_{\text{physical}}(\textbf{x}, \mathcal{F}) = \left[\min(\max(G(\textbf{x}, \mathcal{F}), a), b) - a\right]/(b - a),
\end{equation}
where a physics simulator (or surrogate) \(G(\cdot,\cdot)\), such as a Computational Fluid Dynamics (CFD) solver (\cite{zawawi2018review}), computes the physical performance of the 3D generated artifact \(\textbf{x}\).

\textbf{(b) Domain Constraint with Domain Alignment Measure}: Given a user-defined domain specification \(\mathcal{S}\) and multi-view images \(g(\textbf{x})\) of the 3D object \(\textbf{x}\), we measure the degree of alignment between the object and target domain with the following measure:
\begin{equation} \label{eq:domain}
    \begin{aligned}
        & s_{1} = H_{\text{vlm}}(g(\textbf{x}), \mathcal{S}), & s_{2} = H_{\text{vlm}}(g(\textbf{x}), \neg \mathcal{S}), \\
        & f_{\text{domain}}(g(\textbf{x}), \mathcal{S}, \Gamma) = e^{s_1/\Gamma} / (e^{s_1/\Gamma} + e^{s_2/\Gamma}),
    \end{aligned}    
\end{equation}
where \(\Gamma\) is the temperature hyperparameter that controls the discriminative strength between the positive and negative pairs. Here, we assess the likelihood that a 3D object is within the target domain \(\mathcal{S}\) with a Visual-Language Model \(H_{\text{vlm}}(\cdot,\cdot)\) that sets \(\Gamma=0.01\). More detailed background information is presented in the supplementary section \ref{sec:semantic}.

\textbf{(c) Geometric Novelty Measure}: This measure detects and localizes the regions of geometric novelty by comparing the surface topology of a 3D-generated object with a reference object. The surface topology differences with the reference object indicate regions of geometric novelty that contribute to the improvement or degradation of the overall physical performance \(f_{\text{physical}}\). Here, we instantiate this measure in the context of physical light simulation (\cite{parker2010optix}), capturing the precise surface topology of a 3D object. As such, given a 3D-generated object \(\textbf{x}\) and a target object \(\textbf{y}\) in the conventional set, the measure compares images of the same rendered view, defined as follows: \vspace{-0.5em}
\begin{equation} \label{eq:topo}
    \begin{aligned}
        & s_{3} = \mathop{\sum}_{i=1}^{R} \| \textbf{M} \! \odot \! g(\textbf{y})^{[i]} - \textbf{M} \! \odot \! g(\textbf{x})^{[i]} \|^{2}_{2}, \\
        & s_{4} = \mathop{\sum}_{i=1}^{R}\mathop{\sum}_{j=1}^{J} \| h_{j}(\textbf{M} \! \odot \! g(\textbf{y})^{[i]}) - h_{j}(\textbf{M} \!\odot \! g(\textbf{x})^{[i]}) \|^{2}_{2}, \\
        & f_{\text{novelty}}(\textbf{x}, \textbf{y}) = \frac{\gamma(g(\textbf{x}), g(\textbf{y}))(s_{3}  + s_{4})}{\|\textbf{M}\|_{1}},
    \end{aligned}
\end{equation}
where \(\odot\) denotes the element-wise product, \(R\) is the number of rendered views, \(g(\cdot)^{[i]}\) provides the \(i\)-th rendered view of a 3D object, and \(\textbf{M}\) is a mask matrix for indicating the regions of interest to compare the surface topology. The first term \(s_{3}\) compares the same rendered view image between the 3D-generated and target object at the pixel level. The second term \(s_{4}\) compares various visual feature maps with a pre-trained image feature extractor \(h_{j}(.)\), which extracts specific features at the \(j\)-th level of image abstraction. On a separate note, the measure score is weighted with \(\gamma(\cdot,\cdot)\) to ensure the relevance of the 3D-generated novelty with respect to the target domain. This allows impractical 3D-generated objects that have no resemblance to the target domain to be ranked lower than the practical ones. Details of this weighted function \(\gamma(\cdot,\cdot)\) are discussed in the supplementary section \ref{sec:semantic}. In addition, the \(f_{\text{novelty}}\) is scale invariant by normalizing with the number of pixels considered in the regions of interest \(\|\textbf{M}\|_{1}\). Furthermore, if a set \textbf{Y} of reference objects is provided, the reference object with the least geometric novelty with respect to \textbf{x} is selected as the representative novelty of the set and is defined as: 
\begin{equation} \label{eq:topomin}
    F_{\text{novelty}}(\textbf{x}, \textbf{Y}) = \min_{k \in \{1,...,K\}} \{ f_{\text{novelty}}(\textbf{x}, \textbf{y}_{k}) \, | \, \textbf{y}_{k} \in \textbf{Y}\}.
\end{equation}
\subsection{Meta Prompt Design}
\vspace{-0.5em}
A set of instructions (or meta-prompt) is designed for the LLM to function as an effective optimizer. Specifically, a role description highlights the search task the LLM operates and the goal (or objective) it needs to solve. LLM is provided with target domain knowledge for generating domain-specific prompts, effectively constraining the generation of 3D objects within this domain. To this end, we instruct the LLM to complete the following prompt template, \(\mathcal{M} =\) ``A \(< \mathcal{S} >\) in the shape of'' \label{sec:meta}, where \(\mathcal{S}\) is user specified domain specification. Lastly, for the LLM to improve its generative performance, we provide the LLM with feedback by presenting previously LLM-generated prompts with its evaluated scores as exemplars for the LLM to reflect on and learn from. This in-context learning approach improves the likelihood that the LLM generates more effective prompts than its previous iteration.


\vspace{-1em}
\section{Experiments} \label{sec:exp}
\vspace{-0.5em}
\subsection{Setup and Evaluation} \label{sec:exp_setup}

We demonstrate the efficacy of LLM-to-Phy3D with experiments set up to solve an engineering design optimization problem. The problem is designed to find novel cars with aerodynamically efficient (low drag) body shapes. As such, the target domain \(\mathcal{S}\) is set as ``\textit{Car}'' and we design the meta prompt for the problem with the search goal of minimizing the aerodynamic drag \(\bar{C}_{drag}\) in \(f_{\text{physical}}\) (see Figure \ref{fig:meta} in the supplementary section for an example of the meta prompt). In addition, we bound the drag scores within \(a=-1.0\) and \(b=1.0\) in Equation~\eqref{eq:physical} and normalized. Moreover, we set \(\beta\) accordingly to only reward novel vehicles with better aerodynamic drag performance than a set of reference objects \(\textbf{Y}=\{\textbf{y}_{1},...,\textbf{y}_{Q}\}\). Specifically, we first generate reference objects in \(\textbf{Y}\) with the text-to-3D generative model in the test scope conditioned on ``\textit{A Car}'', and then determine the physical performance of each object. Subsequently, we compute the physical performance standard deviation \(\hat{\sigma}\) and mean \(\hat{\mu}\) of the set \(\textbf{Y}\) for instantiating the hyperparameter as \(\beta = e^{-(\hat{\mu}/\hat{\sigma})}\). We employ CFD physics simulation with OpenFoam (\cite{weller1998tensorial}) to obtain the physical performance aerodynamic scores. In addition, to generate the feature maps in Equation~\eqref{eq:topo}, we utilize the EfficientNet model (\cite{tan2019efficientnet}). Details of how the feature maps are utilized are discussed in the supplementary material Section ~\ref{sec:semantic}. On a separate note, we employ the Nvidia ray tracing engine OPTIX (\cite{parker2010optix}) to capture and render the precise surface topology of 3D objects.

Our proposed method is compared against four different baseline LLM-to-3D methods. We select four LLMs - \textit{GPT-3.5 Turbo}, \textit{GPT-4o Mini}, \textit{Gemini 2.0 Flash Lite}, and \textit{Mistral 3.1 Small}, to compare based on its capabilities in following the designed meta-prompts and creating feasible prompts throughout the iterations in a single experiment run. As these LLMs require a text-to-3D generative model to produce 3D artifacts, we select \textit{Shap-E} and \textit{Trellis} models. For these baseline LLM-to-3D methods, we established a standard meta prompt ``\textit{Create a prompt that starts with \(\mathcal{M}\) and ends with a full stop.}'', where \(\mathcal{M}\) is the prompt template introduced in Section \ref{sec:meta}. Moreover, the temperature of the LLMs is set in a range that allows it to produce viable and diverse prompts. Also, for consistency, we adhere to the same temperature range for these LLMs, unless notable hallucination issues result in a significant degradation of the LLM's ability to follow the instructions. Detailed specifications of these models are recorded in the supplementary section Table~\ref{tbl:llms} and Table~\ref{tbl:3dgens}.
\begin{table}[b]
    \vspace{-1em}
    \caption{Comparison of LLM-to-Phy3D and Baselines \(\textit{DPAR}\) in Producing Physically Conform Target Domain 3D Artifacts} \label{tbl:llms_performance}
    \centering
    \begin{tabular}{l|ccr|ccr}
        \toprule
        \textbf{Text-to-3D Model /} & \multicolumn{3}{c|}{\textbf{Shap-E}} & \multicolumn{3}{c}{\textbf{Trellis}} \\
        \textbf{LLM} & \textbf{Baseline} & \textbf{Ours} & \textbf{Improvement} & \textbf{Baseline} & \textbf{Ours} & \textbf{Improvement} \\
        \midrule
        Mistral 3.1 Small & \(0.8350\) & \(\textbf{0.9557}\) & +14.46\% & \(0.8131\) & \(\textbf{0.8494}\) & +4.46\% \\    
        GPT 3.5 & \(0.8972\) & \(\textbf{0.9637}\) & +7.41\% & \(0.8562\) & \(\textbf{0.9531}\) & +11.32\% \\
        GPT 4o Mini & \(0.6891\) & \(\underline{\textbf{1.0000}}\) & +45.12\% & \(0.6139\) & \(\underline{\textbf{1.0000}}\) & +62.89\% \\
        Gemini 2.0 Lite & \(0.7697\) & \(\textbf{0.9173}\) & +19.18\% & \(0.4660\) & \(\textbf{0.9634}\) & +106.74\% \\
        \bottomrule
    \end{tabular}  
\end{table}
Furthermore, we compare LLM-to-Phy3D against the baseline methods in terms of its ability to produce physically conforming target domain 3D artifacts. Specifically, given the target domain \(\mathcal{S}\) and set \(\boldsymbol{\mathcal{D}}\) of all 3D artifacts generated with a method in a single experiment run, we define the domain and physical alignment rating (\(\textit{DPAR}\)) of a method of interest as:
\begin{equation*} \label{eq:benchmark}
    \textit{DPAR}(\boldsymbol{\mathcal{D}}, \mathcal{S}) = \mathop{\mathbb{E}}_{\textbf{x} \in \mathcal{D}}\left[f_{\text{domain}}\left(g(\textbf{x}), \mathcal{S}\right) / f_{\text{physical}}(\textbf{x})\right],
\end{equation*}
where the higher the rating, the better the method in its ability to produce physically conforming target domain artifacts. For fair comparison, we ensure that the number of generated outputs in a baseline method and our proposed method is the same throughout the iterations (see supplementary section~\ref{sec:parameters} for more details). Furthermore, our proposed method and the baseline use the same LLM temperature ranges.

\vspace{-0.5em}
\subsection{Results and Discussions}
\vspace{-0.5em}
Based on the experiment results, the baseline models often generate vehicles that exhibit aesthetically pleasing shapes or resemble conventional vehicle body shapes (as observed in some of the most representative examples on the left group of Figure~\ref{fig:showcase}). In contrast to these vehicles, LLM-to-Phy3D found novel vehicles that are not only visually creative but also satisfy the physical and target domain constraints (presented in the right group of the same figure). This observation is evident in the benchmark with \(\textit{DPAR}\) (Table~\ref{tbl:llms_performance}), where our proposed method consistently outperforms the baselines by a margin of 4.46\% to 106.74\%. This varying resulting margin of LLM-to-Phy3D is the result of the performance of each LLM in learning the association between the quality of 3D artifacts and its LLM-generated prompts, and predicting the best possible prompts that yield better quality outputs.

\begin{figure}
    \vspace{-2em}
    \centering
    \begin{tabular}{cc}
        \includegraphics[width=0.45\textwidth]{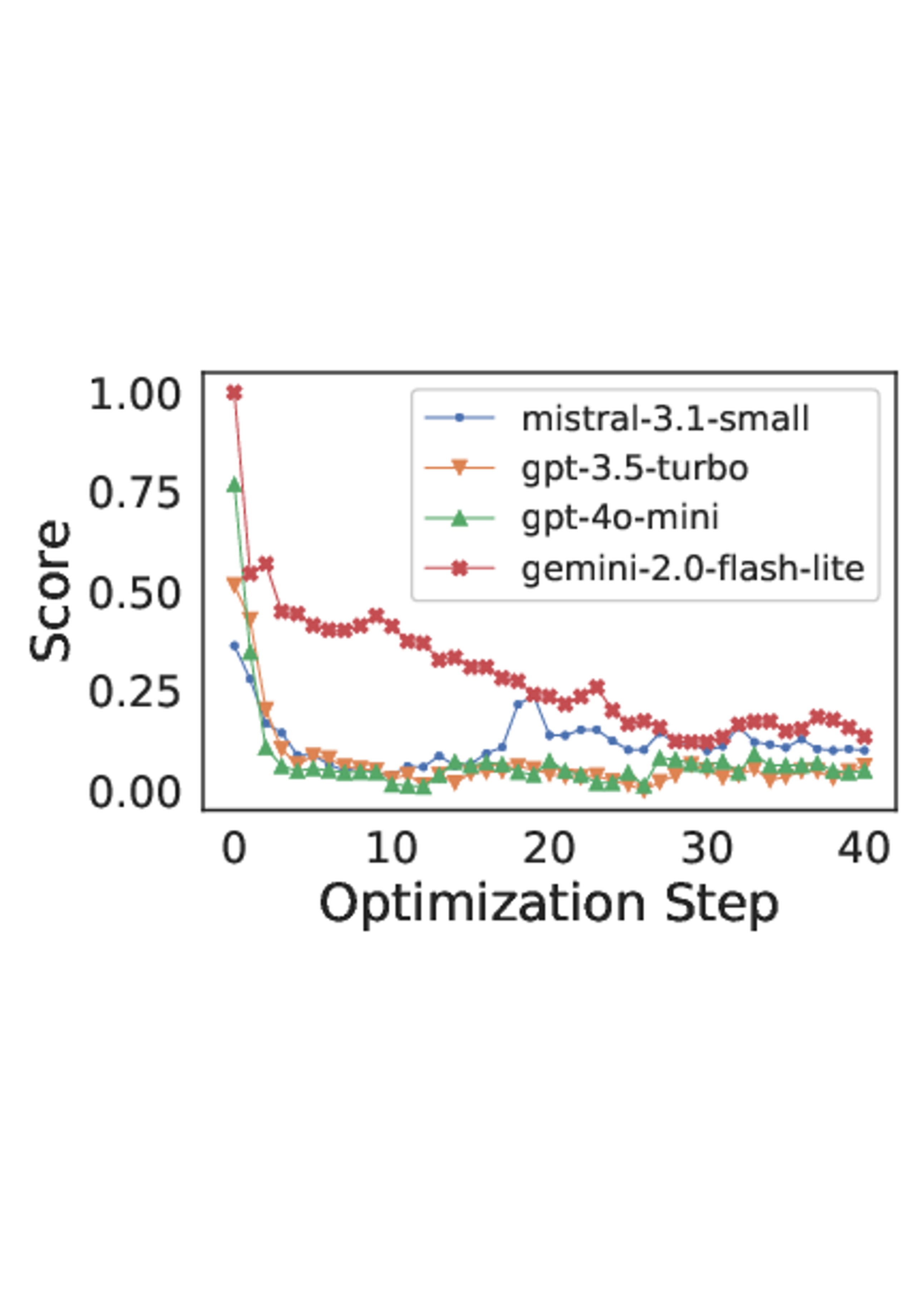} & 
        \includegraphics[width=0.45\textwidth]{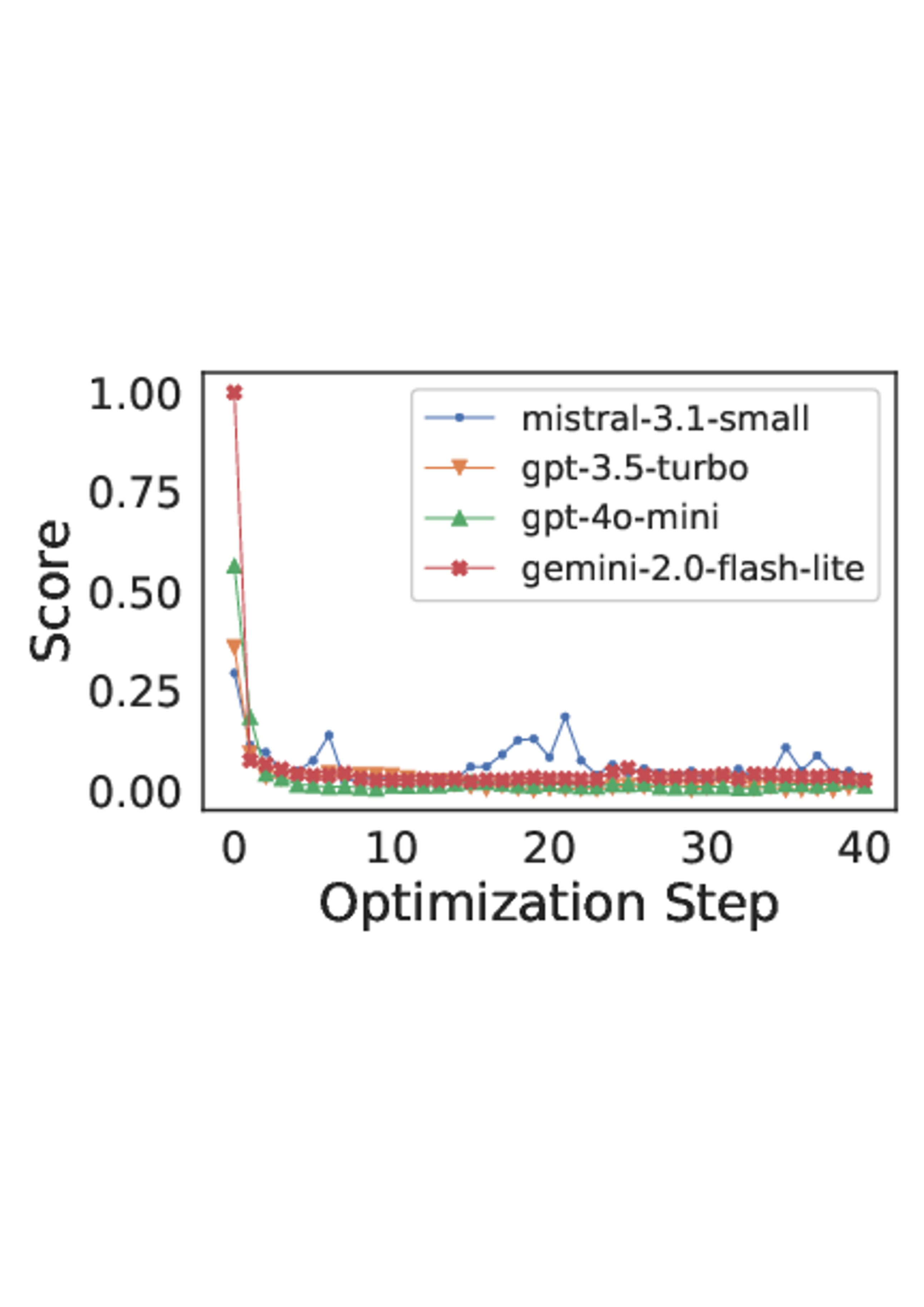} \\ [-70pt]
        \textbf{(a) Shap-E} & \textbf{(b) Trellis} \\ [1pt]
    \end{tabular}
    \caption{Search optimization score (Equation~\ref{eq:obj}) of LLM-to-Phy3D with different LLM and text-to-3D generative models combinations.}
    \label{fig:fitness}
    \vspace{-1em}
\end{figure}

As presented in Figure~\ref{fig:fitness}, \textit{GPT-4o-mini} and \textit{GPT-3.5-Turbo} can learn more proficiently than other LLMs of interest, resulting in convergence in less than 10 steps. On the other hand, while \textit{Gemini 2.0 Flash Lite} is capable of producing diverse and creative prompts, we observe it requires more time to overcome the generation issues in Shap-E (Figure~\ref{fig:fitness}(a)) where the text-to-3D model is less adhering to the LLM-generated prompts. We also noticed that \textit{Mistral 3.1 small} struggles to learn the association between its generated prompt and the evaluated score, hindering its performance in producing effective textual prompts. While still being able to find physically conforming and more novel 3D vehicles, the limitations observed in \textit{Gemini 2.0 Flash Lite} and \textit{Mistral 3.1 small} resulted in minimal variation of the generated prompts.

\begin{figure}
    \centering
    \begin{tabular}{c}
        \includegraphics[width=0.90\textwidth]{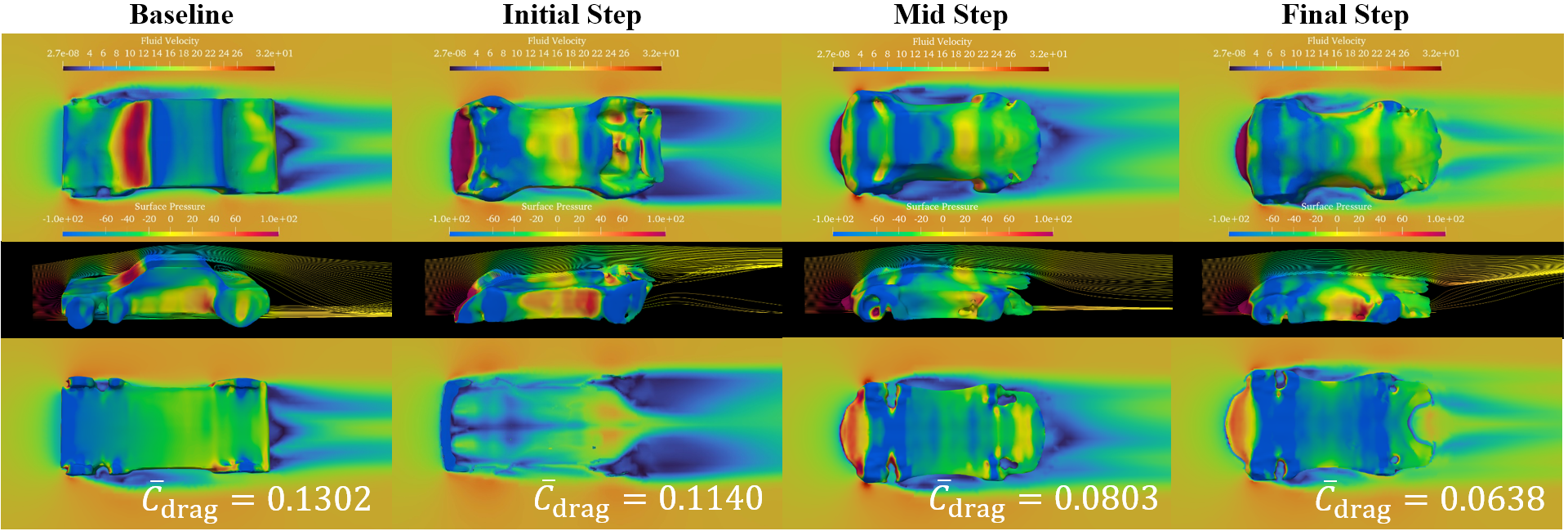} \\
        \textbf{(a) Shap-E} \\ [5pt]
        \includegraphics[width=0.90\textwidth]{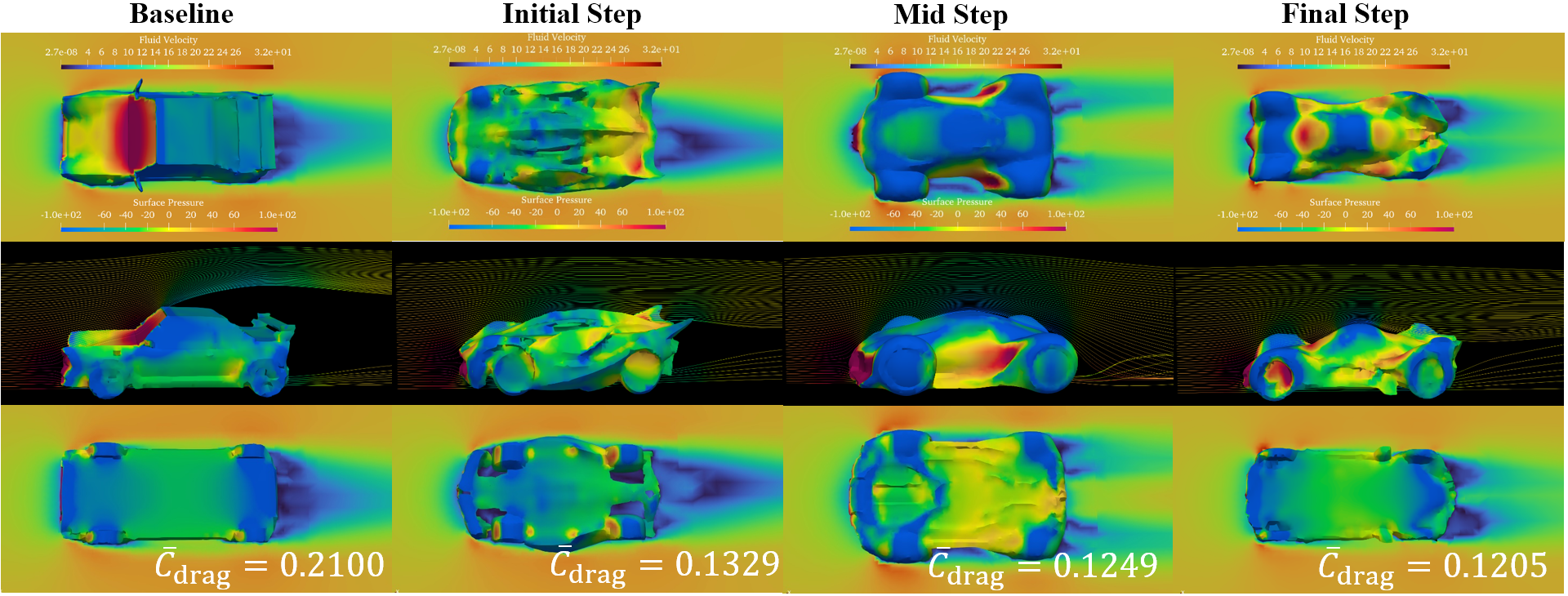} \\
        \textbf{(b) Trellis} \\ [5pt]
    \end{tabular}
    \vspace{-1.5em}
    \caption{Results of the aerodynamic profile of cars generated with LLM-to-Phy3D at specific iterations and compared with a representative car generated with the best baseline LLM-to-3D method. Note that the lower the drag, the better the physical performance of the car.}
    \label{fig:profileflow}
    \vspace{-2em}
\end{figure}

By analyzing the aerodynamic performance of the generated 3D shapes (Figure~\ref{fig:profileflow}), our findings reveal that the initialization step produces 3D-generated cars with shapes that yield high surface pressure at the car front and sides. This increases the aerodynamic drag force and decreases the performance. These resulting artifacts have similar aerodynamic issues to the baseline samples, as the LLMs generate textual prompts without prior knowledge of the physics. After introducing prior knowledge and selecting effective textual prompts that result in better physically conforming 3D artifacts, the generated 3D artifacts possess shapes that lead to lower aerodynamic performance indicators, such as a reduced pressure difference or smaller projected area.

\begin{figure}
    \centering
    \includegraphics[width=1.00\textwidth]{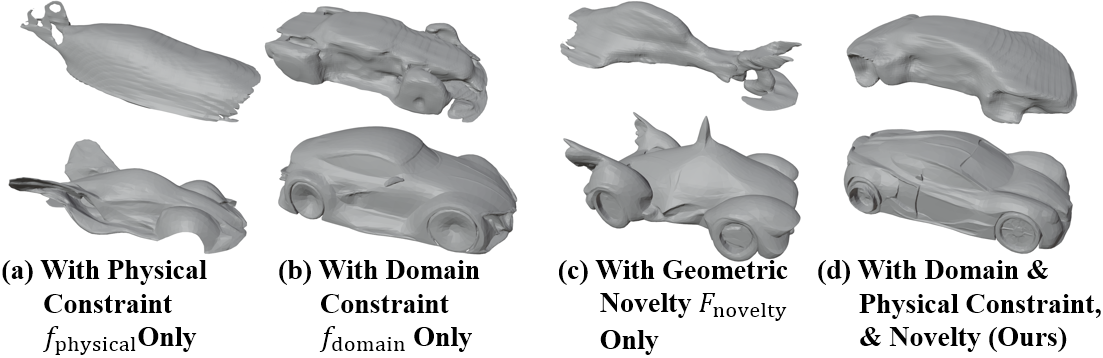}
    \caption{Ablation Studies on the effectiveness of different terms in the objective function on generating 3D artifacts (Top Row: LLM with Shap-E, Bottom Row: LLM with Trellis).}
    \label{fig:ablation}
\end{figure}
\begin{figure}
    \centering
    \includegraphics[width=1.00\textwidth]{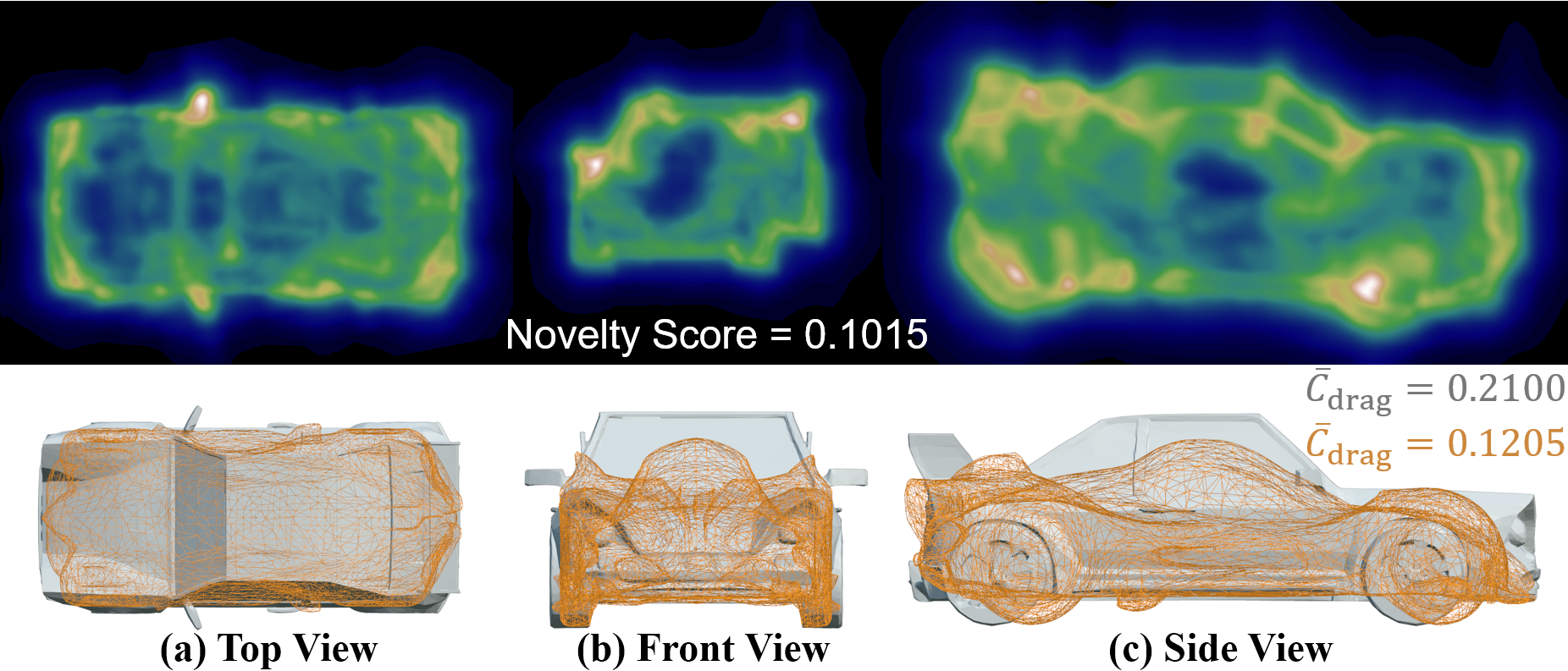}
    \vspace{-0.5em}
    \caption{Example of geometric novelty between a reference car (in {\color[HTML]{B0C5CC}grey} solids) and a car generated with LLM-to-Phy3D (in {\color[HTML]{CD883D}orange} wireframe). The novelty heatmaps presented in the top row are the multi-view results of \(F_{\text{novelty}}(.)\), which highlight the regions contributing to the geometric novelty score of the generated car compared with the target one. Multiple views (bottom row) of the generated car overlay on top of the target car show regions corresponding to the contrasting body shapes attributed to the difference in aerodynamic performance.}
    \label{fig:novelty}
    \vspace{-1em}
\end{figure}

We conducted several ablation studies to examine the effect of each term in the objective function. We noticed that using physical constraint \(f_{\text{physical}}\) only, the search favors malformed or highly fragmented cars (Figure~\ref{fig:ablation} (a)) that registered better drag performance than complete generated cars. This degrades over time the overall performance of the framework as LLM learns the false positives, leading to an incorrect association of prompts with the cars, which results in shapes that yield poorer drag performance. On a separate note, the search process adhering to the target domain constraint \(f_{\text{domain}}\) only can lead to artifacts that visually highly resemble the reference cars of varying physical performance (Figure~\ref{fig:ablation} (b)). With geometric novelty \(F_{\text{novelty}}\) only, on the other hand, LLMs found textual prompts that yield cars with highly unconventional or fragmented shapes that have high novelty but are impractical (Figure~\ref{fig:ablation} (c)). By combining all three terms as a weighted sum in the objective function, LLM-to-Phy3D can effectively produce novel 3D artifacts that satisfy the physical and domain constraints (Figure~\ref{fig:ablation} (d)).

Figure~\ref{fig:novelty} reveals the highlighted visual features indicating the geometric novelty between a reference car and a car generated with LLM-to-Phy3D. In the top row heatmap images in the figure, the geometric differences between the surfaces of a 3D-generated car and a reference one are highlighted, marking the visual cues of high geometric novelty. These differences correlate to the overlay meshes of the cars presented in the bottom row. Such geometric differences provide an indicator of the region of novelty, allowing LLM-to-Phy3D to reward such 3D artifacts in the objective function that not only lead to the discovery of novel artifacts but also satisfy the physical and domain constraints. Investigation on the effectiveness of this geometric novelty measure under different camera projections reveals an LLM-to-Phy3D improvement with orthographic projection (\(\textit{DPAR}~+3.38\%\) to \(+52.21\%\)). This indicates that the geometric distortions due to the projection strategy considerably impact finding physically conforming 3D artifacts.

\vspace{-1em}
\section{Conclusion}
\vspace{-1em}
We introduce \textit{LLM-to-Phy3D}, a novel physically conform text-to-3D online guidance with LLMs. Our proposed method enables existing LLM-to-3D models to produce physically conforming target domain 3D objects on the fly. Central to LLM-to-Phy3D is an online black-box iterative refinement framework that adaptively steers the LLM to find textual prompts that yield physically plausible 3D artifacts with high geometric novelty. Systematic evaluations supported by ablation studies of LLM-to-Phy3D in an aerodynamic vehicle design optimization scenario reveal that our proposed method outperforms existing LLM-to-3D models in producing physically conforming target domain 3D artifacts by up to 106.7\%. Furthermore, the precise surface topology of 3D-generated objects captured through physical light simulation with orthographic projection enables effective measurement of geometric novelty with respect to reference objects. These results underscore the potential general use of LLM-to-Phy3D in Physical AI for further scientific and engineering applications. 

\textbf{Limitations and broader impact:} \label{sec:limitations} Although LLM-to-Phy3D demonstrates remarkable improvements over existing LLMs and text-to-3D models of scope, the pre-existing generation issues and limited expressiveness of these models are inherited, which may limit performance or generate artifacts that are highly fragmented, non-watertight or even containing the multi-face Janus issue (\cite{liu2024direct}). In addition, we envision that our proposed method can accelerate engineering design processes, empowering the labor force with advanced automated AI tools.

\section{Acknowledgements}
Melvin Wong gratefully acknowledges the financial support from Honda Research Institute Europe (HRI-EU). This research is partly supported by the College of Computing \& Data Science (CCDS), Nanyang Technological University (NTU).

\bibliography{references}


\clearpage

\appendix

\section{Technical Appendices and Supplementary Material}

\subsection{Background}


Much interest in computer vision, computer graphics, and continuum dynamics is on the boundary between mediums. In physics, particularly in solid or fluid dynamics, one area of interest is the conditions at the boundary, which are determined by the behavior of particles traveling through the medium as these particles interact with the boundary (\cite{lomax2001fundamentals,shames1997elastic}). For example, the surface of a car defines the physical constraints that govern how particles, such as air, dust, or water, interact with it. On the other hand, in computer vision and computer graphics, a key area of focus is capturing the precise visual features of an object's surface topology (\cite{rothwell1996representing,montagnat2001review,pulli2000surface}). The surface topology provides visual cues to the type of object, its texture, parts of the object, or even occlusion between objects. Such surface topology can be represented by the radiance of light particles as it reflect (and refract) at the object boundary. In this section, we first introduce fluid dynamics, with a primary focus on aerodynamic drag and light under incompressible Newtonian fluid flow. After which, we explore the rendering of surface topology using the ray tracing technique. Lastly, we highlight how the semantic relationship between objects can be measured by comparing the visual features of the objects.

\subsubsection{Fluid Dynamics} \label{sec:drag}

In physics, the fundamental laws and principles governing the behavior of natural physical systems are defined in the governing equations. In fluid dynamics, given the velocity 3D field \(\textbf{u}\), the governing equations for incompressible Newtonian fluids are:

\textbf{Conservation of mass} ensures mass is neither created nor destroyed within the physical system. As such, mass entering a controlled environment equals the mass exiting it. The Continuity equations that describe this behavior in differential form are defined as

\begin{equation} \label{eq:continuity}
    \frac{\partial u_{i}}{\partial x_{i}} = 0,
\end{equation}
where \(u_{i}\) consists of the instantaneous velocity components in all three Cartesian directions. Note that \(i\) is called a \textit{free} index as it is ``free'' to take on any three values in three-dimensional space \((x,y,z)\) (\cite{chaves2013notes}).

\textbf{Conservation of momentum} ensures the rate of change in momentum is the same as the pressure, viscosity, and external forces acting on the fluid particles. The Navier-Stokes equations that describe this relationship as follows:

\begin{equation} \label{eq:navier_stokes}
    \begin{aligned}
        & \sigma_{ij} = -p\delta_{ij} + \mu \left( \frac{\partial u_{i}}{\partial x_{j}} + \frac{\partial u_{j}}{\partial x_{i}} \right), \\
        & \rho\frac{\partial u_{i}}{\partial t} = \frac{\partial \sigma_{ij}}{\partial x_{j}} + \mathcal{F}_{i},
    \end{aligned}
\end{equation}

where \(\delta_{ij}\) is the Kronecker delta, \(\mu\) is the force needed to overcome internal friction in a fluid, \(p\) is the kinematic pressure, \(\rho\) is the fluid density, and \(\mathcal{F}_{i}\) is the body force (e.g. gravity). Here, the Cauchy stress tensor (\(\sigma_{ij}\)) comprises the isotropic pressure, viscous shear, and normal stress. Note that \(i\) and \(j\) are \textit{free} index to denote an axis in three-dimensional space.


\begin{figure}
  \centering
  \includegraphics[width=0.90\linewidth]{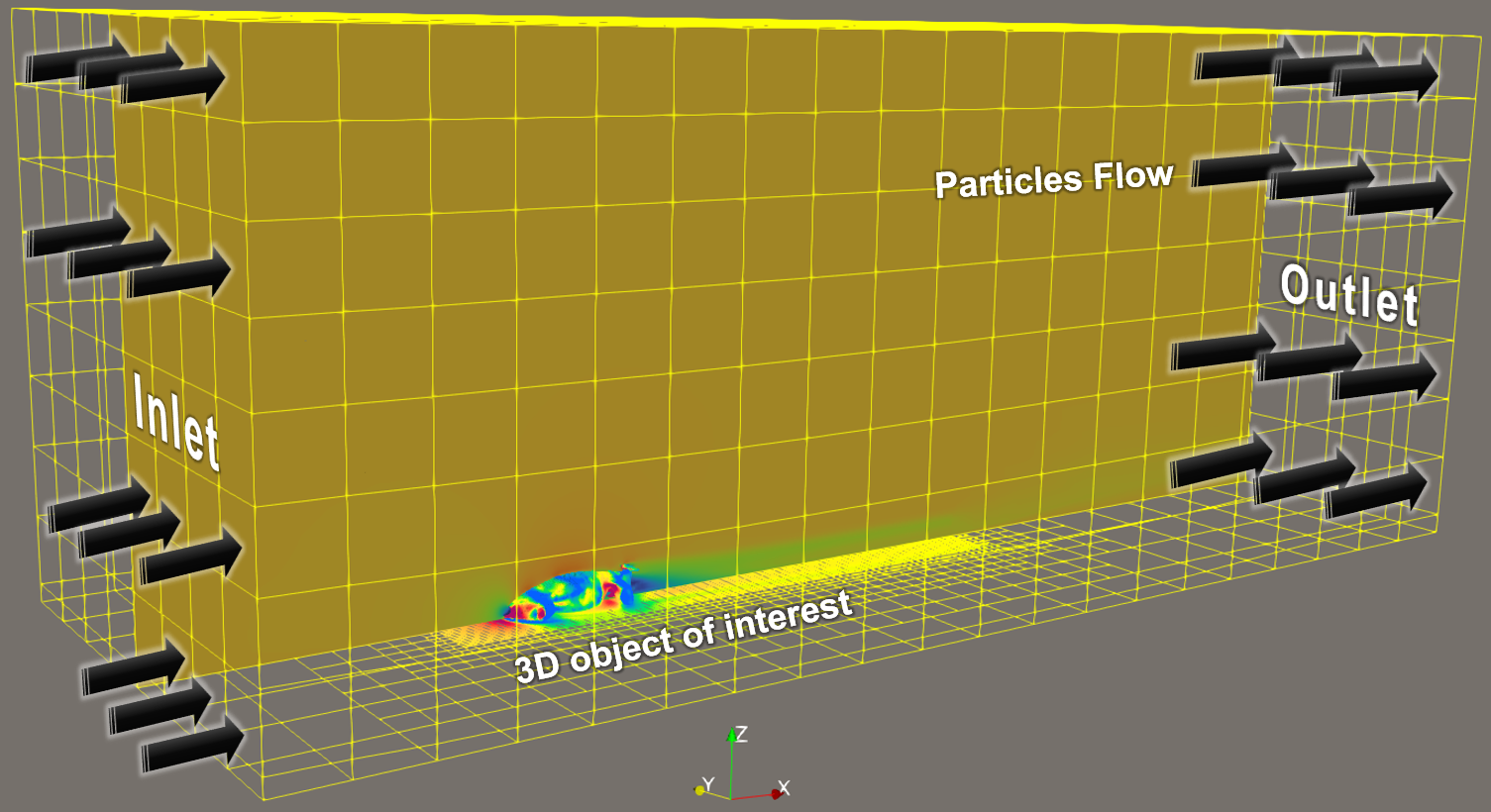}
  \caption{Illustration of virtual wind tunnel for aerodynamics physics simulation. In a simplified controlled environment, the computational domain within the yellow wireframe box simulates the flow of particles from inlet to outlet, as shown in the directional black vector arrows. The physical interactions between the particles and the 3D object of interest are determined by the conditions at the boundary (skin material and surface topology) of the object. These interactions propagated throughout the computational domain form the aerodynamics performance of the 3D object, which can be used as a valid fitness score in a search.}
  \label{fig:cfd}
\end{figure}

To solve the governing equations for the 3D object of interest, the velocity 3D flow is set as the velocity at the surface of the object. As such, given a computational domain \(\Omega\) (Figure~\ref{fig:cfd}), we assume the 3D object of interest \(\textbf{x}\) is located entirely in \(\Omega\), and the spatial regions between \(\textbf{x}\) and the boundary walls of \(\Omega\) are sufficiently large. As uniformly distributed force is applied to displace the Newtonian fluid particles from its stationary state in the direction as shown in Figure~\ref{fig:cfd}, some particles close to the 3D object show different viscous behavior due to physical interactions with the skin and surface of the 3D object.

This viscous behavior is of prime interest as it describes the aerodynamic performance of the 3D object. To quantify the behavior, the pressure on the 3D object is captured to derive the total forces and moments acting on the object. Various aerodynamic properties, such as lift, drag, and pressure, are then computed. In this paper, the drag aerodynamic property is our main focus.

Without loss of generality, given a 3D object \(\textbf{x}\), the aerodynamic forces acting on the body of the object are:

\begin{equation} \label{eq:fluid_forces}
    \begin{aligned}
        & \mathcal{F}_{\text{normal\_pressure}}^{[i]} = \left[p - p_{\text{reference}}(\textbf{x})\right]S_{\text{normal}}^{[i]}, \\
        & \mathcal{F}_{\text{viscous}}^{[i]} = \tau_{\text{wall}}^{[i]} = d_{\infty} (v - v_{t})\left[\frac{\partial u_{i}}{\partial x_{j}} + \frac{\partial u_{j}}{\partial x_{i}}\right]S_{\text{normal}}^{[j]}(\textbf{x}), \\
        & \mathcal{F}_{\text{total}}^{[i]} = \mathcal{F}_{\text{normal\_pressure}}^{[i]} + \mathcal{F}_{\text{viscous}}^{[i]},
    \end{aligned}
\end{equation}

where \(S_{\text{normal}}^{[i]}\) is the surface normal vector, and \(\tau_{\text{wall}}^{[i]}\) is the wall shear stress force vector. Note that \(i\) and \(j\) are \textit{free} index to denote an axis in three-dimensional space. The total forces in non-dimensional form are defined as follows:

\begin{equation} \label{eq:forces}
    \mathcal{F}_{xyz} = \int_{S} \mathcal{F}_{\text{total}}^{[i]} \textbf{N}_{xyz}^{[i]} \, dS,
\end{equation}

where \(S\) is the surface points, and \(\textbf{N}_{xyz}^{[i]}\) is the unit vectors pointing in the \(x\), \(y\), and \(z\) directions. Note that \(i\) is a \textit{free} index to denote an axis in three-dimensional space.



\subsubsection{Surface Topology Rendering With Ray Tracing and Global Illumination} \label{sec:rendering}

A ray is cast from the camera viewpoint to the center of a pixel in the image and towards infinity. The ray is defined as follows:

\begin{equation} \label{eq:ray}
    r(t) = \textbf{o} + t \cdot \textbf{d},
\end{equation}

where \(\textbf{o}\) is the origin of the ray, \(t\) is the distance from the ray origin, and \(\textbf{d}\) is the normalized direction vector of the ray. If the ray hits a surface at an intersection point \(\textbf{z}\), the direction of ray reflection is:

\begin{equation} \label{eq:ray_reflection}
    \textbf{d}_{\text{reflection}} = \textbf{d} - 2 (\textbf{d} \cdot \textbf{n})\textbf{n},
\end{equation}

where \(\textbf{n}\) is the normalized surface normal vector at \(\textbf{z}\). On the other hand, the direction of ray refraction is:

\begin{equation} \label{eq:ray_refraction}
    \begin{aligned}
        & \cos{\theta} = -(\textbf{n} \cdot \textbf{d}), \\
        & \lambda = 1 - \eta^{2} \left(1 - \cos^{2}{\theta} \right), \\
        & \textbf{d}_{\text{refraction}} = \eta\textbf{n} + \left(\cos{\theta} - \sqrt{\lambda}\right)\textbf{n}, \quad \text{if} \,\, \lambda \geq 0,
    \end{aligned}
\end{equation}

where \(\eta = \eta_{\text{incident}} / \eta_{\text{transmitted}}\) is the relative refractive index of the incident medium over the transmitted medium (refractive indexes). In this paper, we assume total internal reflection occurs and hence, no refraction ray is produced (\(\lambda < 0\)).

Given the surface of a 3D object \(\textbf{x}\) is represented in triangular meshes where each triangle has vertices \(\textbf{v}_{1}, \textbf{v}_{2}, \textbf{v}_{3}\), the edges of a triangular mesh can be defined as:

\begin{equation} \label{eq:edges}
    \begin{aligned}
        & \textbf{e}_{1} = \textbf{v}_{2} - t \textbf{v}_{1}, \\
        & \textbf{e}_{2} = \textbf{v}_{3} - t \textbf{v}_{1}.
    \end{aligned}    
\end{equation}

The intersection of a ray with a triangular mesh is determined by checking:

\begin{equation} \label{eq:intersection}
    \begin{aligned}
        & \textbf{h} = \textbf{d} \times \textbf{e}_{2}, \\
        & \Lambda = \textbf{e}_{1} \cdot \textbf{h}, \\
        & \textbf{s} = \textbf{o} + \textbf{v}_{1}, \\
        & u = \frac{\textbf{s} \cdot \textbf{h}}{\Lambda}, \\
        & v = \frac{\textbf{d} \cdot (\textbf{s} \times \textbf{e}_{1})}{\Lambda}, \\
        & t = \frac{\textbf{e}_{2} \cdot (\textbf{s} \times \textbf{e}_{1})}{\Lambda}, \\
    \end{aligned}    
\end{equation}

where \(u\) and \(v\) are the barycentric coordinates of the intersection point relative to the triangle. Notice that no intersection occurs when \(a\) is infinitesimally small. Otherwise, if \(0 \leq u \leq 1\), \(0 \leq v \leq 1\), and \(t > 0\), the ray falls within the triangular edges and onto the mesh, indicating an intersection is occurring. In this scenario, under global illumination, the intensity (or shading) at the intersection point \(\textbf{z}\) is defined as follows:

\begin{equation} \label{eq:illumination}
    I(\textbf{z}) = I_{\text{direct}}(\textbf{z}) + I_{\text{indirect}}(\textbf{z}).
\end{equation}

Assuming the surface of a 3D object reflects light equally in all directions under the Lambertian reflection model, the direct illumination at \(\textbf{z}\) is given by

\begin{equation} \label{eq:direct_illumination}
    I_{\text{direct}}(\textbf{z}) = \mathop{\sum}_{i=1}^{L} m_{\text{diffuse}} \cdot \max(0, \textbf{n} \cdot \textbf{L}_{i}) \cdot V(\textbf{z}, \textbf{L}_{i}),
\end{equation}

where \(m_{\text{diffuse}}\) is the diffuse reflection coefficient (material color), \(\textbf{L}_{i}\) is the normalized light direction vector for the \(i\)-th light source, \(\textbf{n}\) is the normalized surface normal vector at \(\textbf{z}\), and \(V(.)\) is the visibility function that returns a scalar value indicating the amount of light reaching \(\textbf{z}\). On the other hand, the indirect illumination accounts for light that has bounced off other surface(s) before reaching \(\textbf{z}\) and is given by:

\begin{equation} \label{eq:indirect_illumination}
    I_{\text{indirect}}(\textbf{z}) = m_{\text{diffuse}} \mathop{\int}_{\Omega} I_{\text{incoming}}(\textbf{z}, \textbf{d}_{i}) \cdot \max(0, \textbf{n} \cdot \textbf{d}_{i}) \, d\textbf{d}_{i},
\end{equation}

where \(m_{\text{diffuse}}\) is the diffuse reflectance of the surface, \(\Omega\) is the hemisphere around \(\textbf{z}\), \(\textbf{d}_{i}\) is a unit vector representing an incoming light direction, and \(I_{incoming}(\textbf{z}, \textbf{d}_{i})\) is the incoming light intensity from direction \(\textbf{d}_{i}\). As this indirect illumination is analytically insolvable, it is normally approximated with Monte Carlo integration (\cite{dutre2018advanced}), given as:

\begin{equation} \label{eq:approx_indirect_illumination}
    I_{\text{indirect}}(\textbf{z}) \approx m_{\text{diffuse}} \frac{1}{N} \mathop{\sum}_{j=1}^{N} I_{\text{incoming}}(\textbf{z}, \textbf{d}_{i}) \cdot \max(0, \textbf{n} \cdot \textbf{d}_{i}),
\end{equation}

where \(N\) is the number of sampled directions. Indirect illumination can distort the surface topology due to the blending of multiple light sources, which may alter visual features and lead to false positives in geometric novelty when comparing object shapes. In overcoming such issues, we consider multiple views of the object during comparisons to minimize the error caused by such distortion.

\begin{figure}
  \centering
  \includegraphics[width=1.00\linewidth]{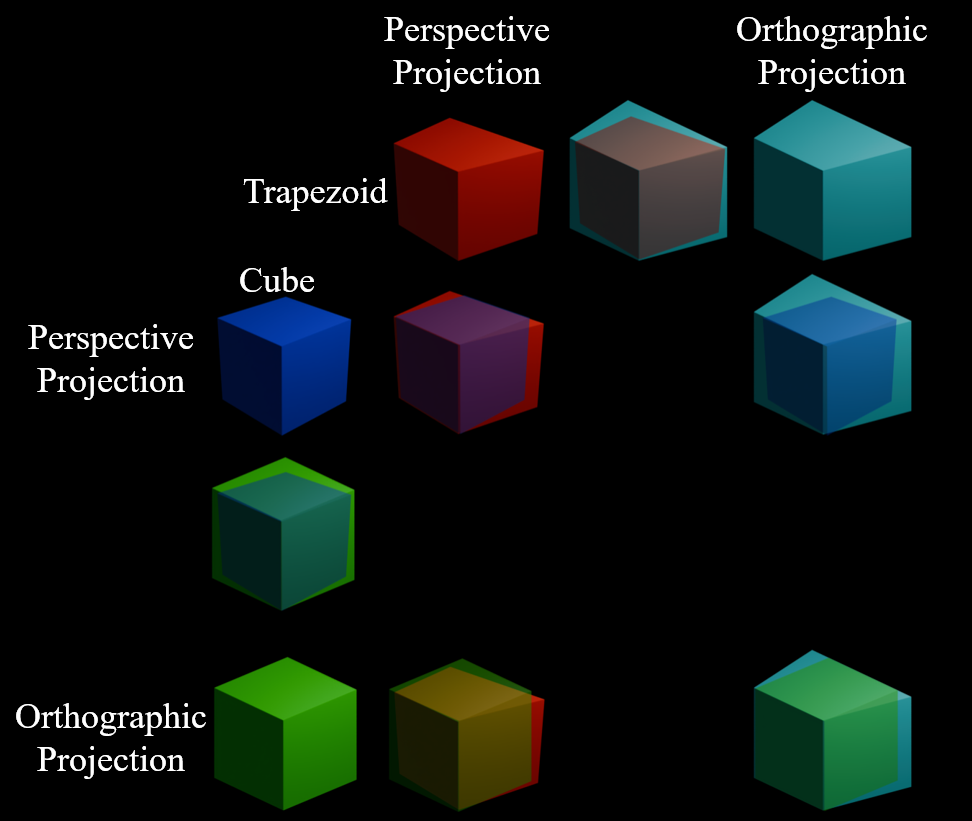}
  \caption{Example of objects rendered under different camera projections and overlay projected objects under the same and different projections to demonstrate the geometric distortions. As the perspective projection does not preserve dimensions and ratios, geometric distortions occur, particularly at the far end of objects, which can cause a violation of the ordering of geometric novelty between the target object and the generated 3D object.}
  \label{fig:distortion}
  \vspace{-1em}
\end{figure}

Furthermore, projecting 3D objects onto a 2D plane can cause geometric distortions. Perspective projection is commonly used in digital art as it mimics how the human eye perceives depth, bringing visual realism to digital media applications (\cite{xiao2022facial,Bolkart_2023_CVPR,Teepe_2024_CVPR,Dutta2025CHROMECH,Zhang2024Diff3DSGV}). However, such a projection does not preserve the geometric dimensions and ratios, particularly at the far end of the objects, as parallel lines converge to a vanishing point (see Figure ~\ref{fig:distortion}). This makes it inapplicable for comparing the geometric differences between objects of varying shapes. In contrast, orthographic projection projects objects onto a plane with parallel lines that are perpendicular to the plane, allowing it to preserve the distance and object shape. Such a projection is suitable for engineering design applications. Hence, in this paper, we conduct a physical light simulation using the ray tracing technique under orthographic camera projection to further minimize geometric distortions.

\subsubsection{Measuring Semantic Relationship} \label{sec:semantic}

The semantic relationship between concepts or entities provides meaningful correlation information about the subjects of interest. One of the effective ways to quantify such information is to use Vision-Language Models (VLMs). VLMs are trained on the visual and textual information of different objects to learn the underlying domain distributions, connecting visual cues of objects with common keywords. This unified representation provides an effective and convenient way to measure the semantic relationships between objects. In this paper, we focus on measuring the semantic relationships between instances of entities as well as entities to a concept. Specifically, given a set \(\mathbb{X}\) consisting of \(R\) view images of a 3D generated artifact, we measure the semantic relationship of the latter as follows:

\begin{equation} \label{eq:image2text}
    H_{\text{vlm}}(\mathbb{X}, \mathcal{S})=\max _{i \in \{1,...,R\}} \Biggl\{\max\left(0,\frac{E_{\text{image}}(\textbf{X}^{[i]}) \cdot E_{\text{text}}(\mathcal{S})^{\text{T}}}{\|E_{\text{image}}(\textbf{X}^{[i]})\|\|E_{\text{text}}(\mathcal{S})^{\text{T}}\|}\right)\Biggr\},
\end{equation}


where \(\textbf{X}^{[i]} \in \mathbb{X} \, \forall i \in \{1,...,R\}\), and \(E_{\text{image}}\) and \(E_{\text{text}}\) are embedding models from pre-trained Vision-Language Model (VLM). In contrast, measuring the semantic relationship between instances of the same or different entities is defined as:
\vspace{-0.1em}
\begin{equation} \label{eq:image2image}
    H_{\text{vm}}(\textbf{X}, \textbf{Y}) = \frac{E_{\text{image}}(\textbf{X}) \cdot E_{\text{image}}(\textbf{Y})^{\text{T}}}{\|E_{\text{image}}(\textbf{X})\|\|E_{\text{image}}(\textbf{Y})^{\text{T}}\|},
\end{equation}

where \(\textbf{X}\) and \(\textbf{Y}\) are images of different entity instances in the same rendered view. We visually assess whether object \(\textbf{x}\) is similar to another object \(\textbf{Y}\) by comparing the rendered views as follows: 

\begin{equation} \label{eq:semantic}
    \begin{aligned}
    & \forall i \in \{1,...,R\}, \, \textbf{X}^{[i]} \in \mathbb{X}, \, \textbf{Y}^{[i]} \in \boldsymbol{\mathcal{Y}}, \\
    & \gamma(\mathbb{X}, \boldsymbol{\mathcal{Y}}) = \frac{1}{R}\left(\sum_{i=1}^{R} \left[\max\left(0, H_{\text{vm}}(\textbf{X}^{[i]}, \textbf{Y}^{[i]}) \right)\right]\right),
    \end{aligned}
\end{equation}

where comparison from \(R\) multiple views minimizes false positives due to high similarity in low-level visual features (e.g. lines, edges, or texture patterns), which may be caused by the surface topology distortions during rendering. In this paper, we use the BLIP2 VLM model (\cite{li2023blip}) to measure the semantic relationship.

\begin{table}
  \caption{Overview of Text-to-3D Generation Methods}
  \centering
  \begin{tabular}{l|c|c}
    \toprule
    & \multirow{2}{7em}{\centering\textbf{Generalize to Target Domain}} & \multirow{2}{7em}{\centering\textbf{Conform to Physics}} \\
    \textbf{Method} & & \\
    \midrule
    \textbf{Diffusion-SDF} \citet{li2023diffusion}  & \cmark & \xmark \\
    \textbf{Shap-E} \citet{jun2023shap} & \cmark & \xmark \\
    \textbf{Trellis} \citet{xiang2024structured} & \cmark & \xmark \\
    \textbf{LLaMA-Mesh} \citet{wang2024llama} & \cmark & \xmark \\
    \midrule
    \textbf{Fun3D} \citet{xu2025fun3d} & \xmark & \cmark \\
    \citet{rios2023large} & \xmark & \cmark \\
    \textbf{PREDO} \citet{wong2024prompt} & \xmark & \cmark \\
    \midrule
    \textbf{LLM-to-Phy3D (Ours)} & \cmark & \cmark \\
    \bottomrule
  \end{tabular}
  \label{tbl:methods}
\end{table}

\begin{figure}
  \centering
  \includegraphics[width=1.00\linewidth]{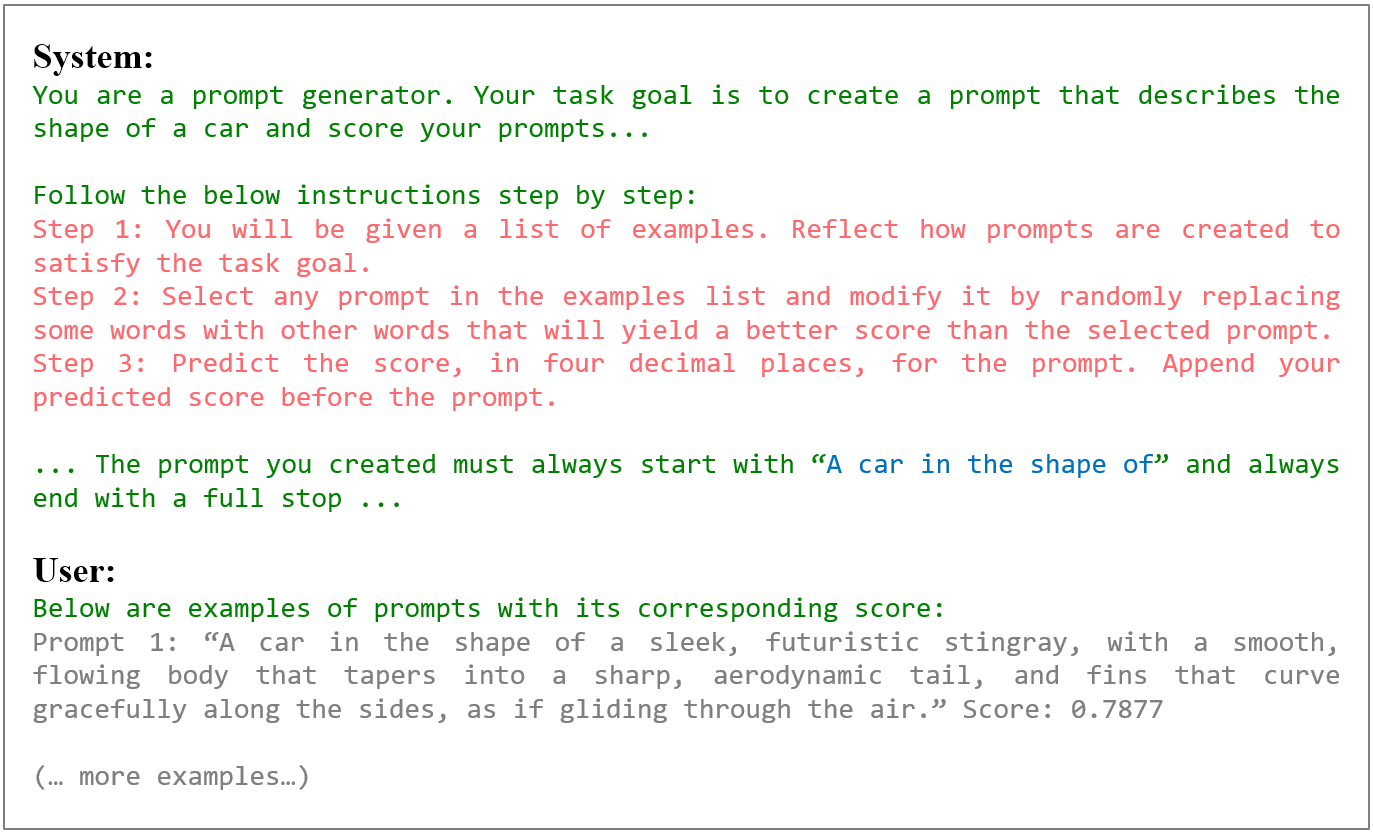}
  \caption{Example of meta prompt designed for aerodynamic design optimization task. The instructions highlighted in {\color[HTML]{008000}green} provide contextual information on the role, the search task the LLM needs to solve, and the operating constraints it must follow. On the other hand, the steps (in {\color[HTML]{FC696F}pink} provide the intermediate sequential operations LLM needs to follow strictly for producing new prompt candidates. Moreover, LLM is instructed to embed the domain specification in the prompts it generates, following the prefix provided in {\color[HTML]{0070C0}blue}. Every iteration after initialization will provide good exemplars of its previous attempt (as shown in {\color[HTML]{7F7F7F}gray}) to guide LLM in producing better prompts.}
  \label{fig:meta}
\end{figure}

\subsection{Failure Scenarios and Mitgation Approaches} \label{sec:failure}
The inherited generation issues from the base LLM and text-to-3D models due to hallucination will lead to LLM producing incoherent outputs. Our proposed method detects this case in LLM by verifying that the generated prompt still contains the prompt template \(\mathcal{M}\) and ends with a full stop. For generated prompts that failed to pass this verification, LLM will re-attempt the generation. 

Moreover, the text-to-3D model will at times produce 3D artifacts with non-watertight surfaces, resulting in incomplete boundaries. While some artifacts are repairable with Trimesh (\cite{trimesh}), text-to-3D model will generate another 3D artifact with the same LLM-generated prompt for non-repairable ones. We also observe some generated artifacts with shapes that do not conform to physics, resulting in the physical simulation reporting evaluation scores with extremely large or small values. Henceforth, we bound such scores within an acceptable range, with such non-conforming artifacts taking the extreme bounded value, which will eventually be filtered out in the selection process. In addition, OpenFoam will sometimes pre-terminate the simulation for such non-conforming artifacts by throwing an error. Our proposed method will handle this exception by regenerating another 3D artifact with the same LLM-generated prompt.

\subsection{Experiment Parameter Settings and Visualization} \label{sec:parameters}

In all experiments, the number of candidates generated in a step is set to $N=20$, and the same random seeds are used throughout the iterations. We ran each experiment for \(40\) search steps, resulting in up to 800 designs generated in a single experiment run. Furthermore, in a single experiment run, the experiment is performed on a single shared compute node with a configuration comprised of Intel Xeon Silver 64 CPU cores clocked at 2.10 GHz, 128GB of RAM, and three Nvidia Quadro GV100 GPUs (32 GB each). For visualizing the aerodynamic drag performance of the generated 3D cars, particularly the surface pressure and fluid velocity, we use ParaView (\cite{fabian2011paraview}) for visualization.


\begin{table}
  \caption{Benchmark Large Language Models} \label{tbl:llms}
  \centering
  \begin{tabular}{l|l|c|r|r}
    \toprule
    \textbf{Provider} & \textbf{Model Name} & \textbf{\# Params} & \textbf{Cut-Off Knowledge} & \textbf{Temperature}\\
    \midrule
    OpenAI & GPT 3.5 Turbo & undisclosed & Sep 2021 & 0.90 to 1.30 \\
    OpenAI & GPT 4o Mini & undisclosed & Oct 2023 & 0.90 to 1.30 \\    
    Google & Gemini 2.0 Flash Lite & undisclosed & Jun 2024 & 0.15 to 1.00 \\
    Mistral & Mistral 3.1 Small & 24B & \(\sim\)Oct 2023 (\cite{allmoComprehensiveList}) & 0.15 to 1.00 \\
    \bottomrule
  \end{tabular}
\end{table}

\begin{table} 
  \caption{Benchmark 3D Generative Models} \label{tbl:3dgens}
  \centering
  \begin{tabular}{l|l|l|c}
    \toprule
    \textbf{Model Name/Variant} & \textbf{Reference} & \textbf{Type} & \textbf{No. Of Params} \\
    \midrule
    Shap-E & \citet{jun2023shap} & Text-to-3D & 300M \\
    Trellis Text Large & \citet{xiang2024structured} & Text-to-3D & 1.1B \\
    \bottomrule
  \end{tabular}
\end{table}

\subsection{Additional Experiment Results and Discussions}
\begin{figure}[b]
  \centering
  \includegraphics[width=1.00\linewidth]{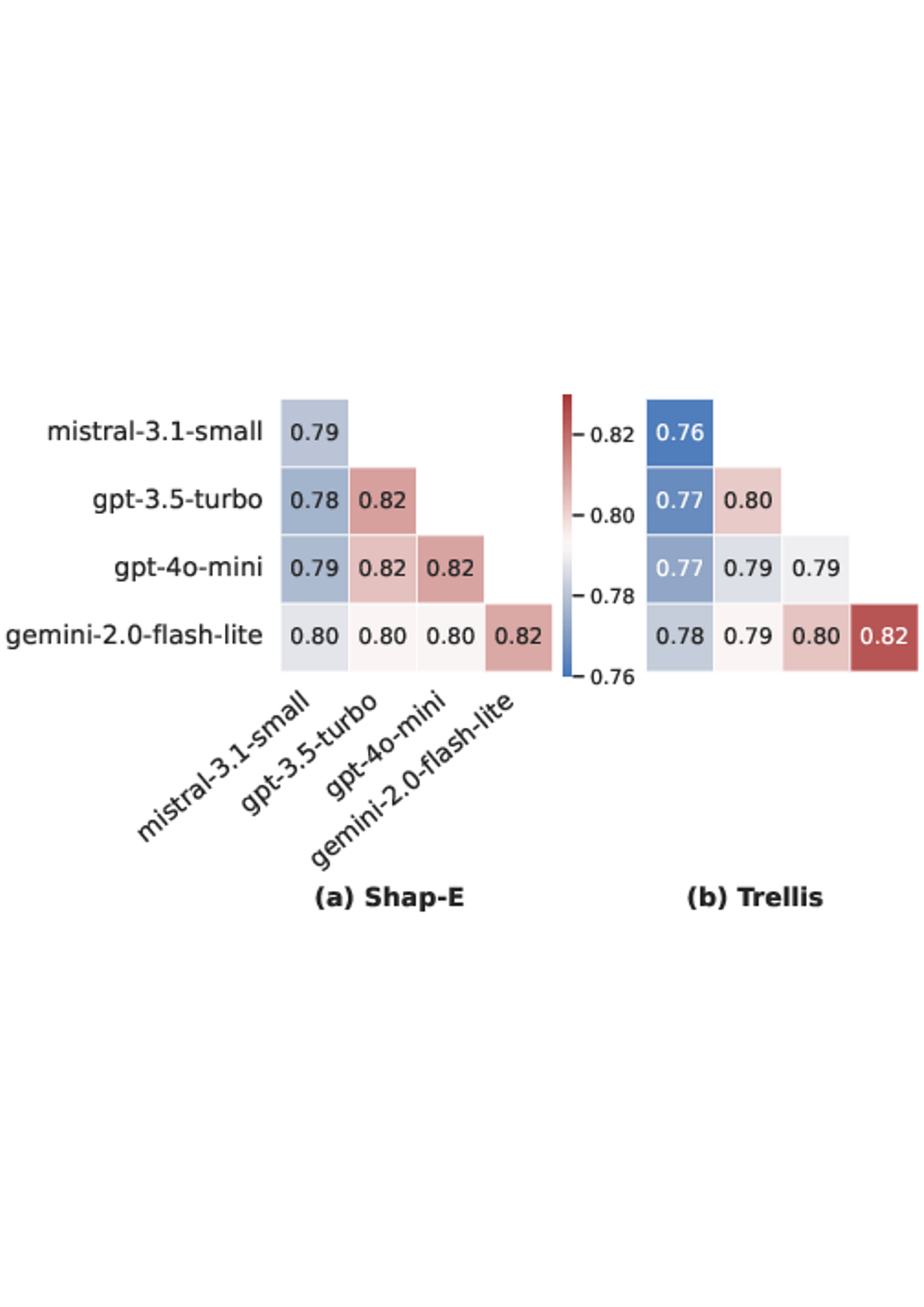}
  \caption{Similarity of 3D generated artifacts between and in various LLMs in the last iteration.}
  \label{fig:corr}
  \vspace{-1em}
\end{figure}

Interestingly, we observed \textit{Mistral 3.1 small} produces the set of 3D artifacts at the final iteration that is less similar to the ones synthesized with other LLMs (Figure~\ref{fig:corr}). This suggests a divergence of the LLMs in mapping the association between the textual and 3D information, leading to different exploration behavior that discovers the varying 3D artifacts.

\begin{figure}
  \centering
  \begin{tabular}{cc}
    \includegraphics[width=0.50\linewidth]{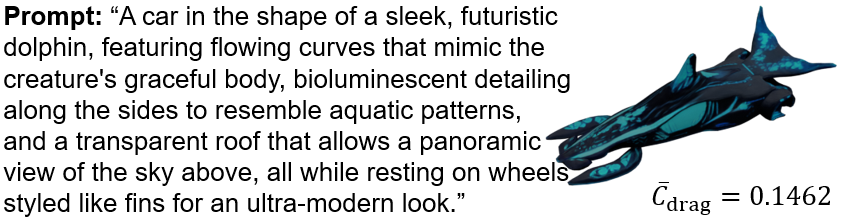} & 
    \includegraphics[width=0.50\linewidth]{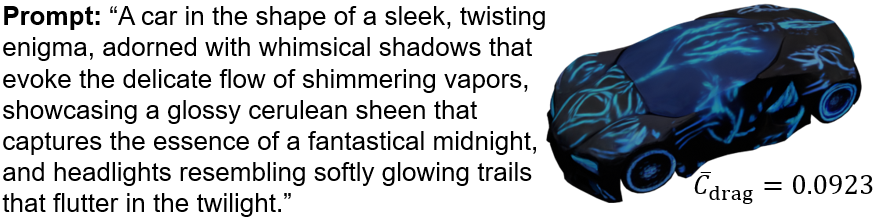} \\
    \includegraphics[width=0.50\linewidth]{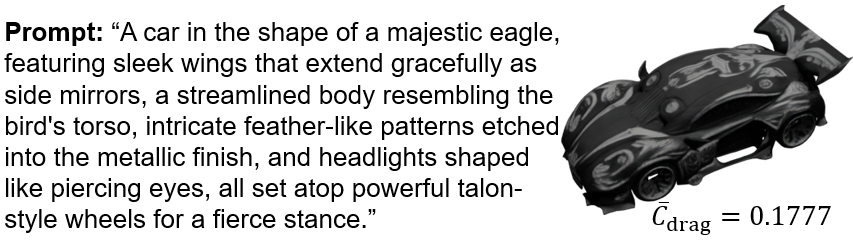} & 
    \includegraphics[width=0.50\linewidth]{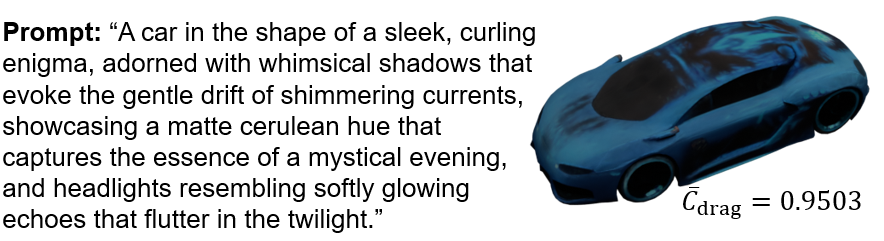} \\
    \includegraphics[width=0.50\linewidth]{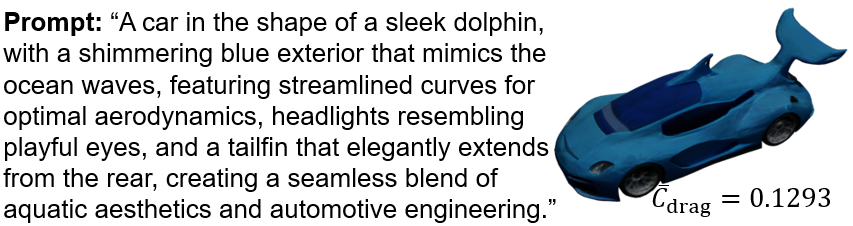} & 
    \includegraphics[width=0.50\linewidth]{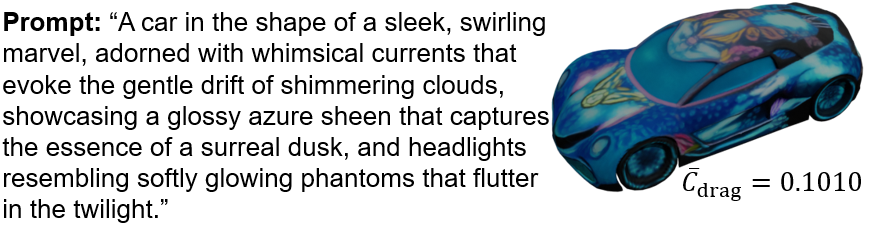} \\
    \includegraphics[width=0.50\linewidth]{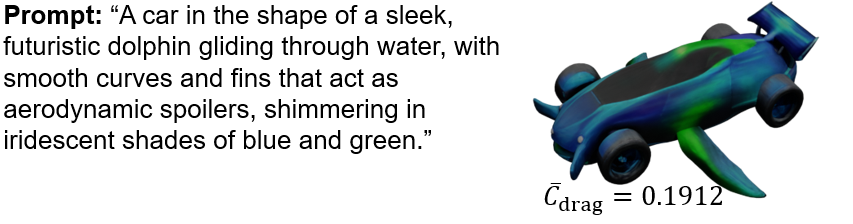} & 
    \includegraphics[width=0.50\linewidth]{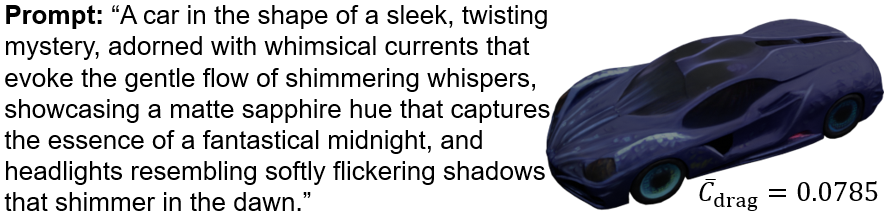} \\
    \includegraphics[width=0.50\linewidth]{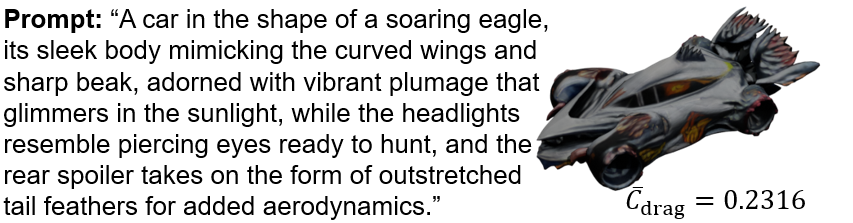} & 
    \includegraphics[width=0.50\linewidth]{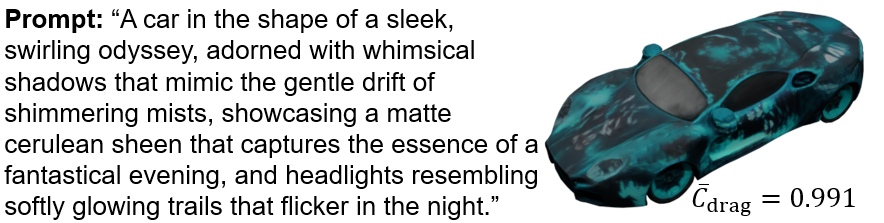} \\
    \includegraphics[width=0.50\linewidth]{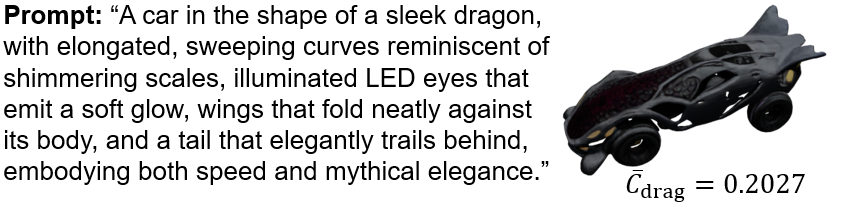} & 
    \includegraphics[width=0.50\linewidth]{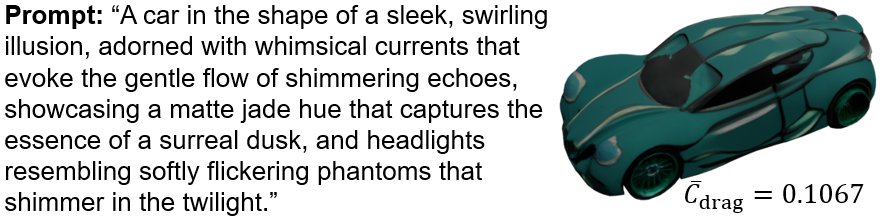} \\
    \includegraphics[width=0.50\linewidth]{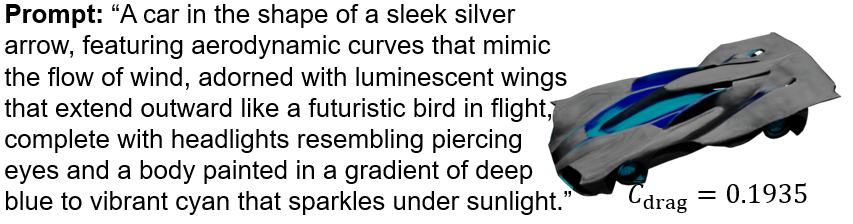} & 
    \includegraphics[width=0.50\linewidth]{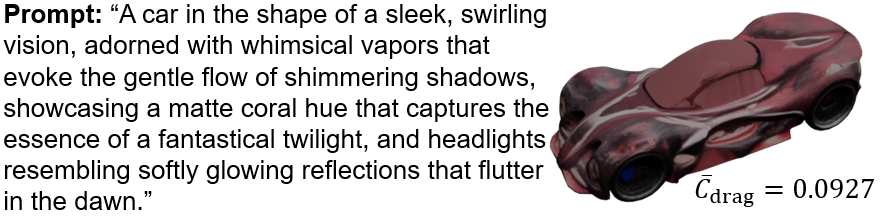} \\
    \includegraphics[width=0.50\linewidth]{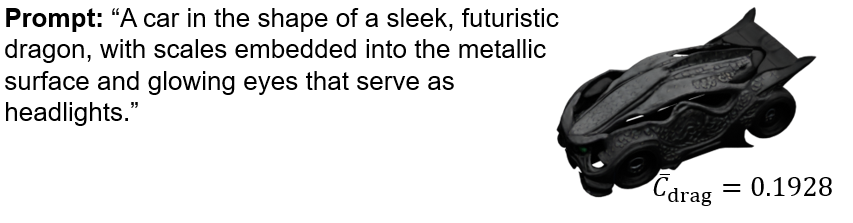} & 
    \includegraphics[width=0.50\linewidth]{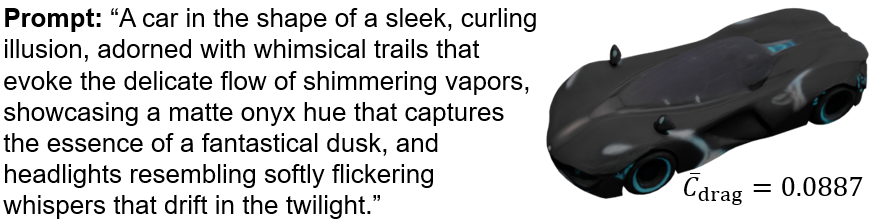} \\
    \includegraphics[width=0.50\linewidth]{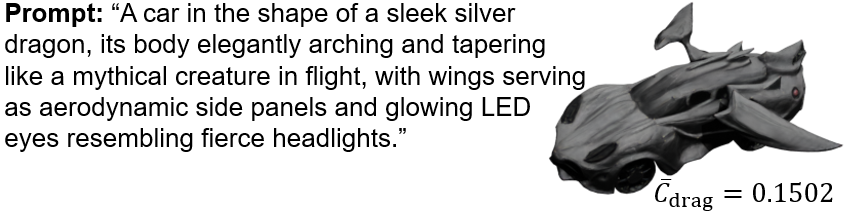} & 
    \includegraphics[width=0.50\linewidth]{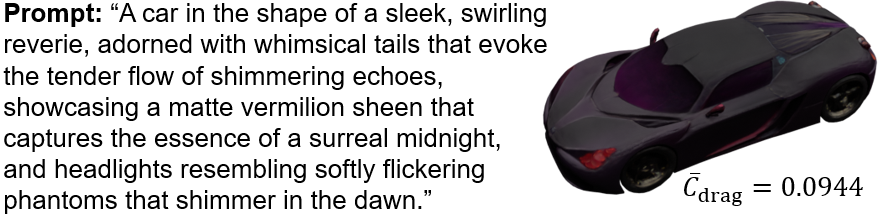} \\
    \textbf{(a) Existing GPT-4o-Mini with Trellis model} & \textbf{(b) LLM-to-Phy3D} \\ [5pt]
  \end{tabular}
  \caption{Examples of 3D cars generated with existing GPT-4o-Mini and Trellis text-to-3D generative model (Left) and with LLM-to-Phy3D (Right). Note that the lower the aerodynamic drag, the better the physical performance of the generated car.}
\end{figure}

\begin{figure}
  \centering
  \begin{tabular}{cc}
    \includegraphics[width=0.50\linewidth]{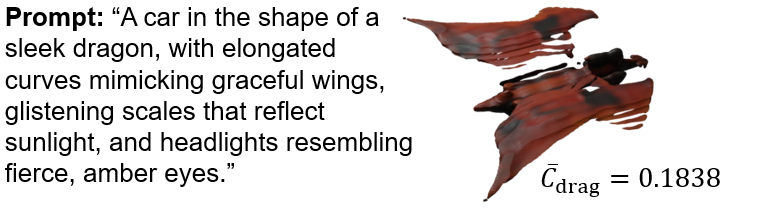} & 
    \includegraphics[width=0.50\linewidth]{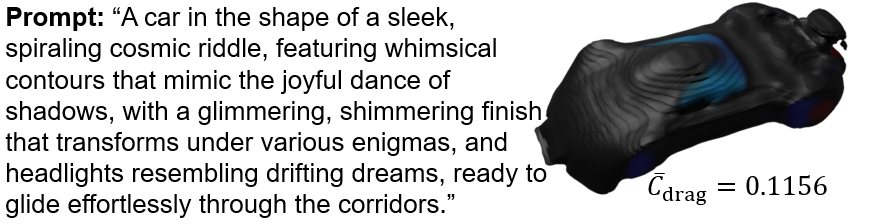} \\
    \includegraphics[width=0.50\linewidth]{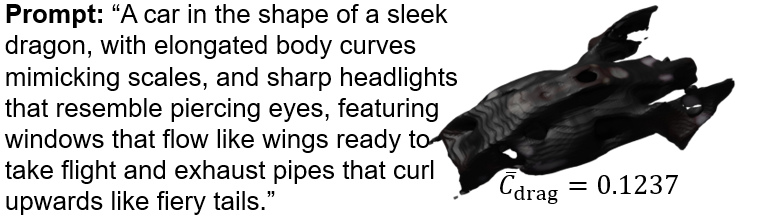} & 
    \includegraphics[width=0.50\linewidth]{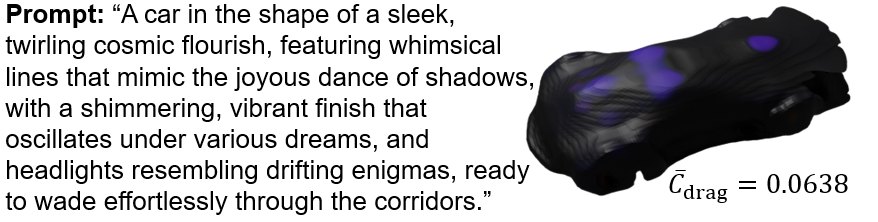} \\
    \includegraphics[width=0.50\linewidth]{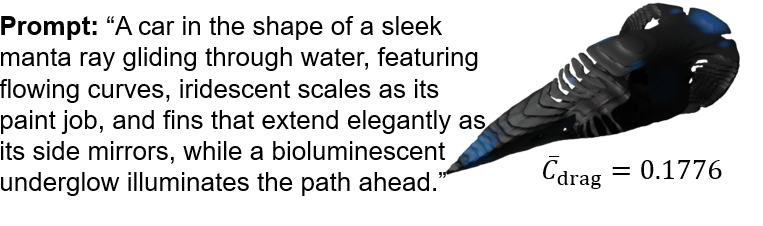} & 
    \includegraphics[width=0.50\linewidth]{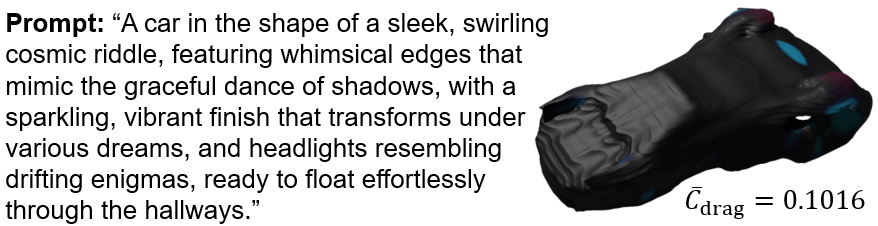} \\
    \includegraphics[width=0.50\linewidth]{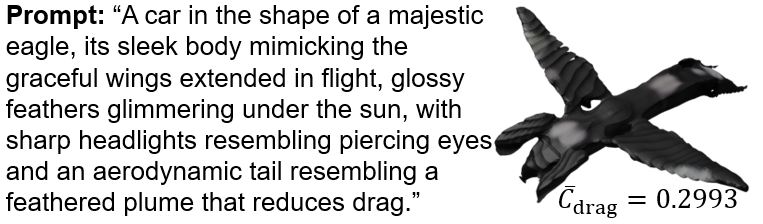} & 
    \includegraphics[width=0.50\linewidth]{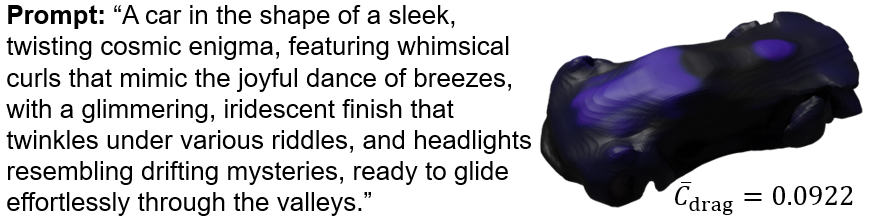} \\
    \includegraphics[width=0.50\linewidth]{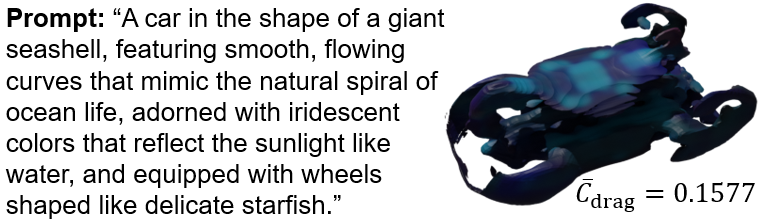} & 
    \includegraphics[width=0.50\linewidth]{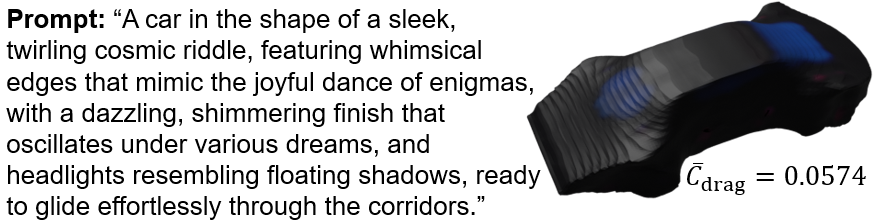} \\
    \includegraphics[width=0.50\linewidth]{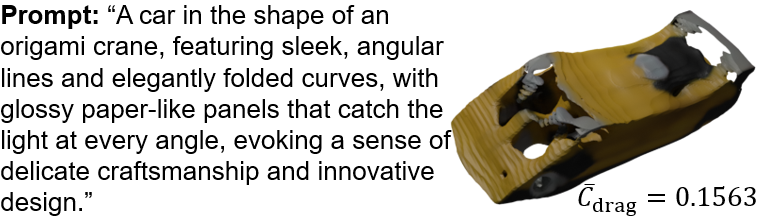} & 
    \includegraphics[width=0.50\linewidth]{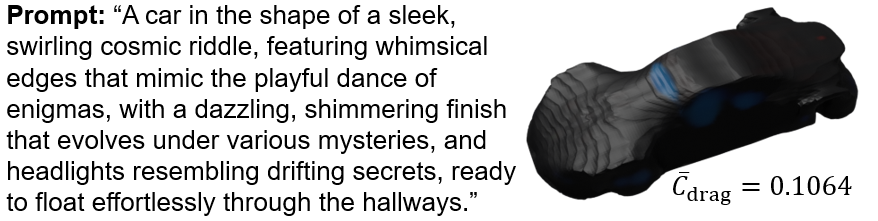} \\
    \includegraphics[width=0.50\linewidth]{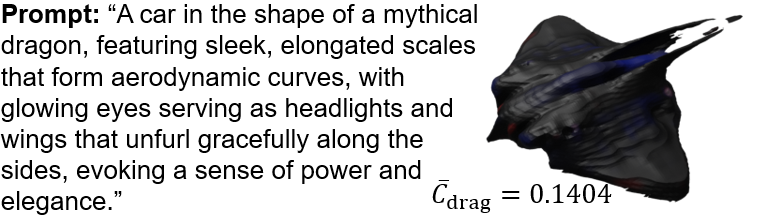} & 
    \includegraphics[width=0.50\linewidth]{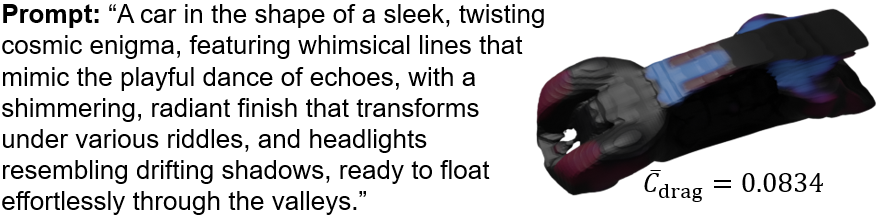} \\
    \includegraphics[width=0.50\linewidth]{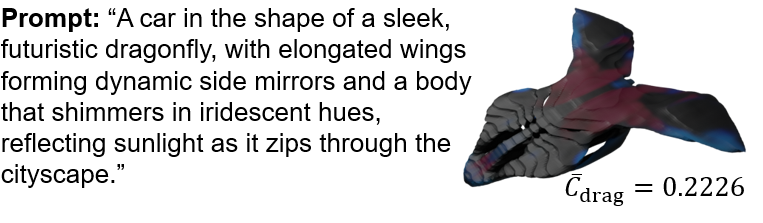} & 
    \includegraphics[width=0.50\linewidth]{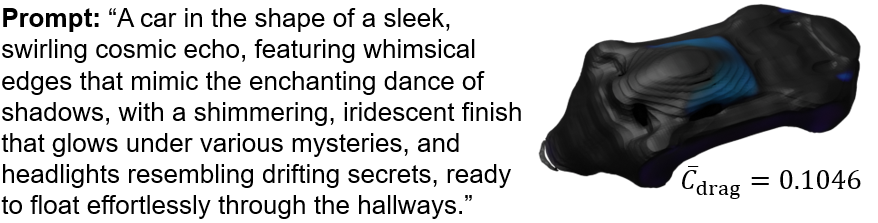} \\
    \includegraphics[width=0.50\linewidth]{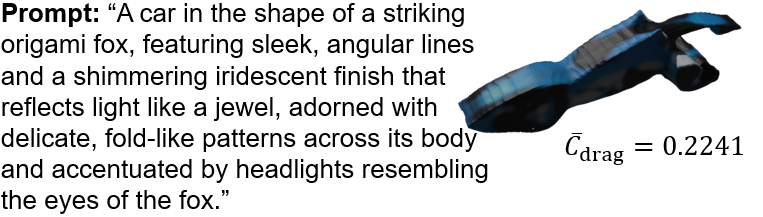} & 
    \includegraphics[width=0.50\linewidth]{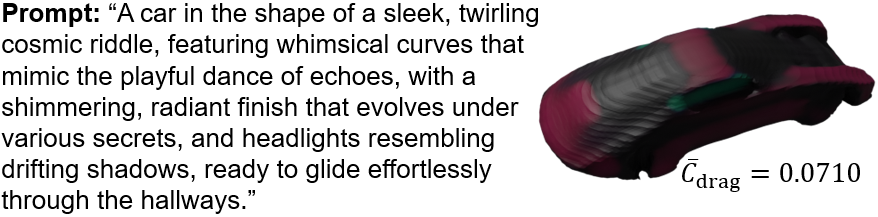} \\
    \textbf{(a) Existing GPT-4o-Mini with Shap-E model} & \textbf{(b) LLM-to-Phy3D} \\ [5pt]
  \end{tabular}
  \caption{Examples of 3D cars generated with existing GPT-4o-Mini and Shap-E text-to-3D generative model (Left) and with LLM-to-Phy3D (Right). Note that the lower the aerodynamic drag, the better the physical performance of the generated car.}
\end{figure}

\begin{figure}
  \centering
  \begin{tabular}{cc}
    \includegraphics[width=0.48\linewidth]{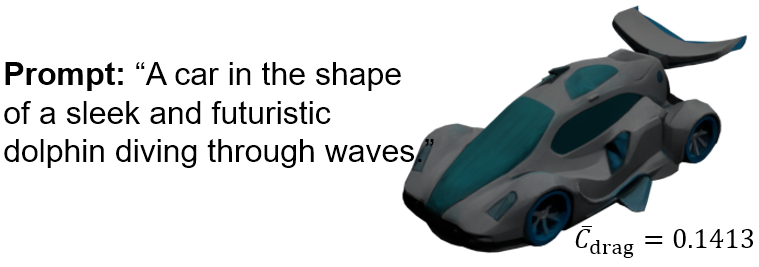} & 
    \includegraphics[width=0.48\linewidth]{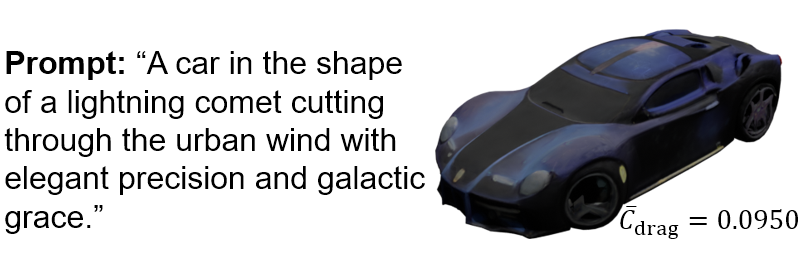} \\
    \includegraphics[width=0.48\linewidth]{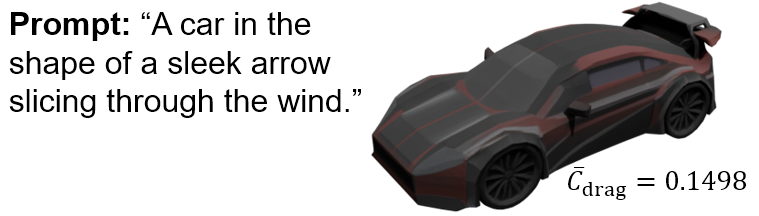} & 
    \includegraphics[width=0.48\linewidth]{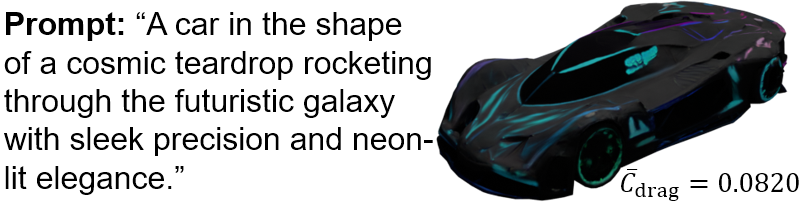} \\
    \includegraphics[width=0.48\linewidth]{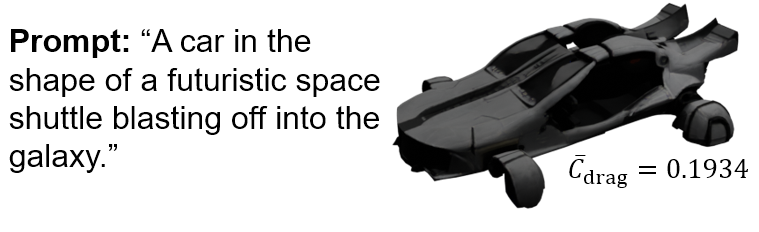} & 
    \includegraphics[width=0.48\linewidth]{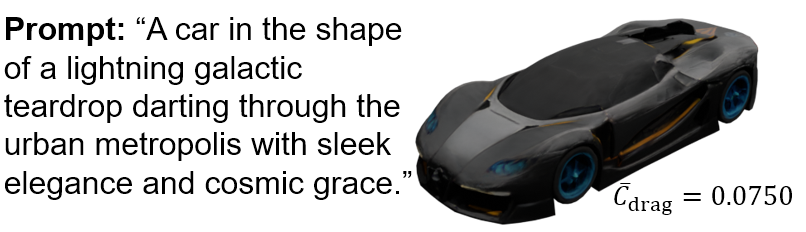} \\
    \includegraphics[width=0.48\linewidth]{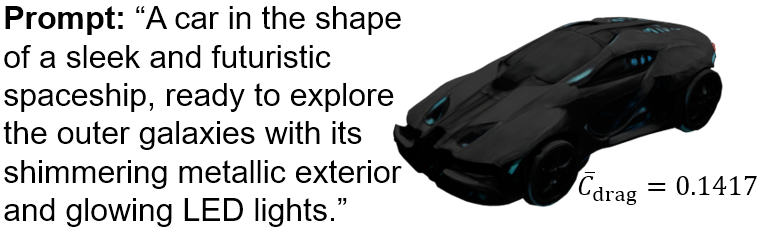} & 
    \includegraphics[width=0.48\linewidth]{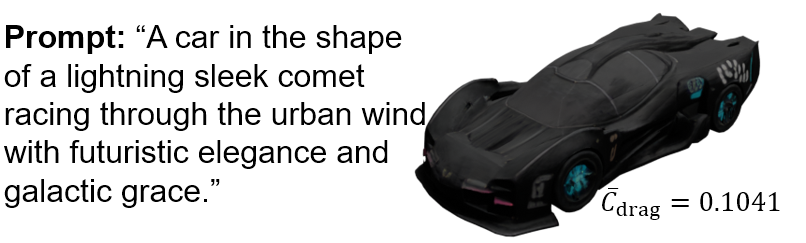} \\
    \includegraphics[width=0.48\linewidth]{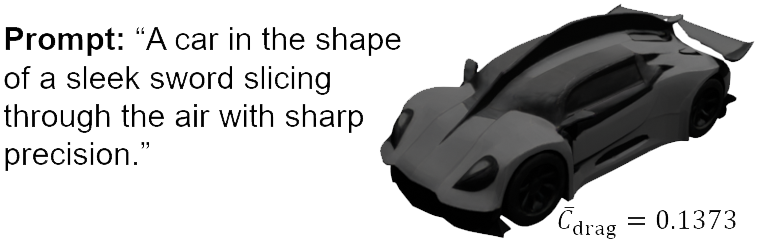} & 
    \includegraphics[width=0.48\linewidth]{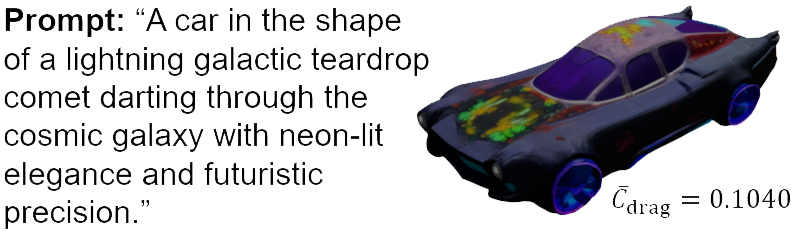} \\
    \includegraphics[width=0.48\linewidth]{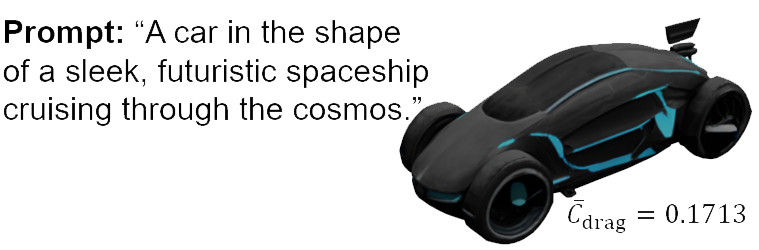} & 
    \includegraphics[width=0.48\linewidth]{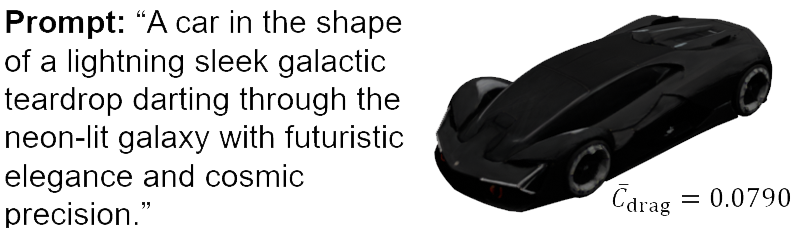} \\
    \includegraphics[width=0.48\linewidth]{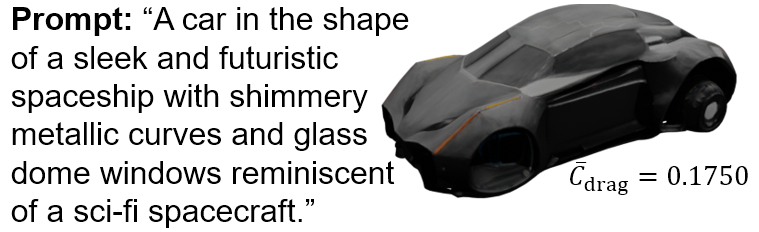} & 
    \includegraphics[width=0.48\linewidth]{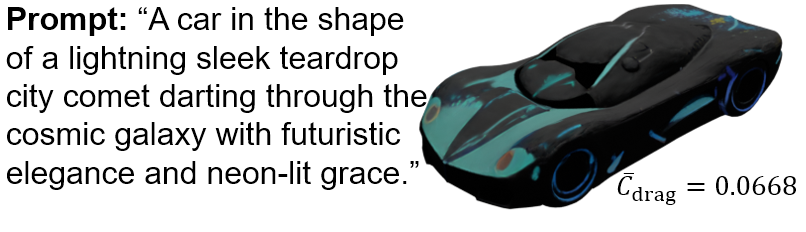} \\
    \includegraphics[width=0.48\linewidth]{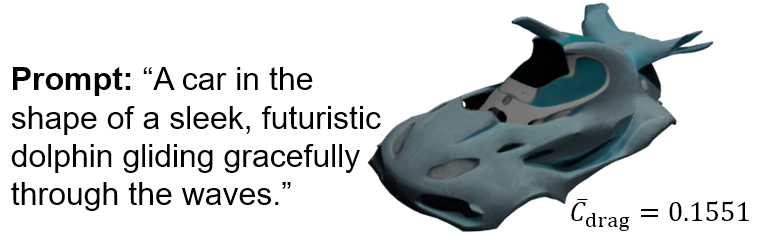} & 
    \includegraphics[width=0.48\linewidth]{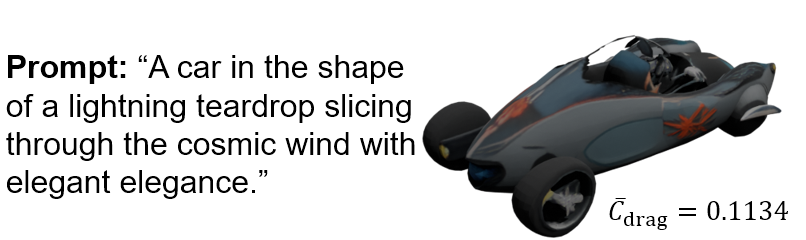} \\
    \includegraphics[width=0.48\linewidth]{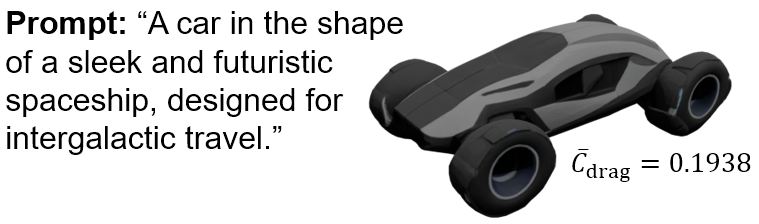} & 
    \includegraphics[width=0.48\linewidth]{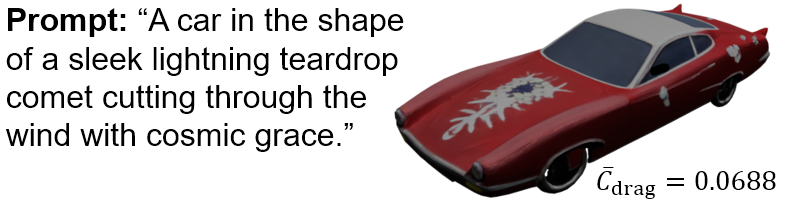} \\
    \textbf{(a) Existing GPT-3.5-Turbo with Trellis model} & \textbf{(b) LLM-to-Phy3D} \\ [5pt]
  \end{tabular}
  \caption{Examples of 3D cars generated with existing GPT-3.5-Turbo and Trellis text-to-3D generative model (Left) and with LLM-to-Phy3D (Right). Note that the lower the aerodynamic drag, the better the physical performance of the generated car.}
\end{figure}

\begin{figure}
  \centering
  \begin{tabular}{cc}
    \includegraphics[width=0.48\linewidth]{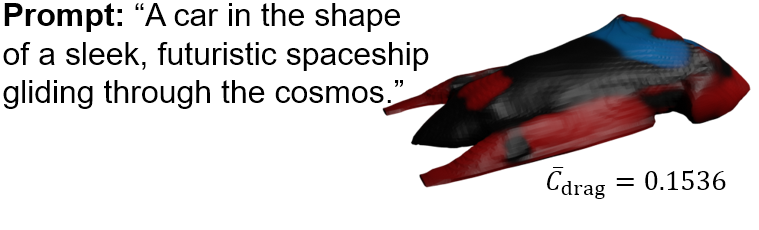} & 
    \includegraphics[width=0.48\linewidth]{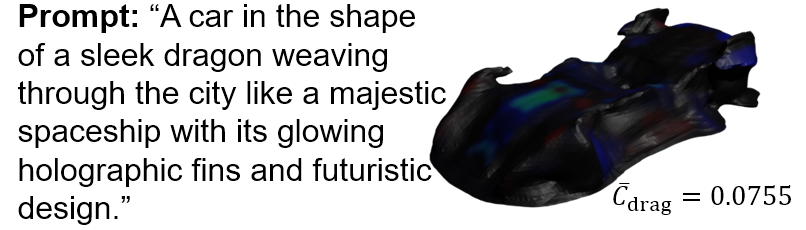} \\
    \includegraphics[width=0.48\linewidth]{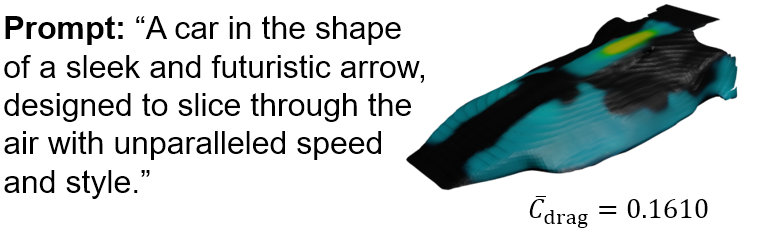} & 
    \includegraphics[width=0.48\linewidth]{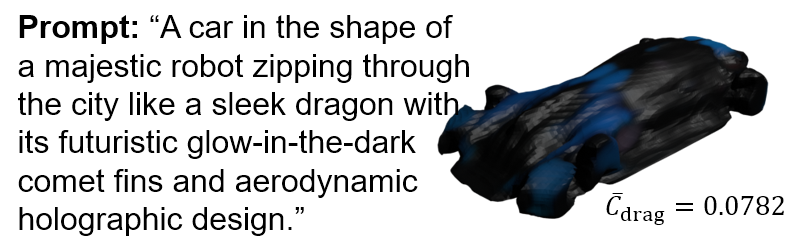} \\
    \includegraphics[width=0.48\linewidth]{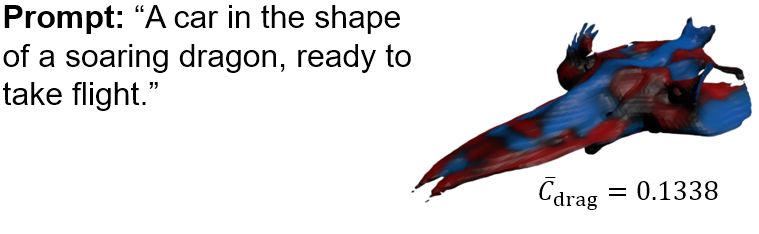} & 
    \includegraphics[width=0.48\linewidth]{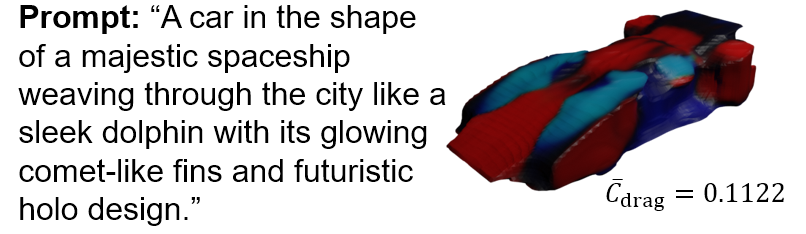} \\
    \includegraphics[width=0.48\linewidth]{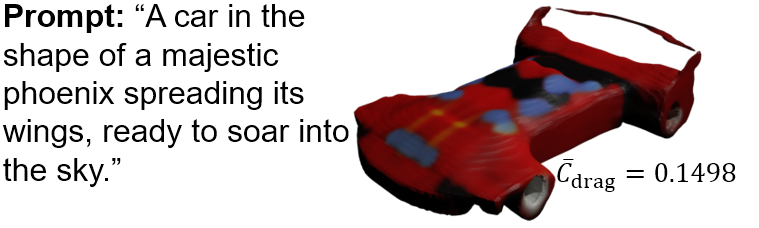} & 
    \includegraphics[width=0.48\linewidth]{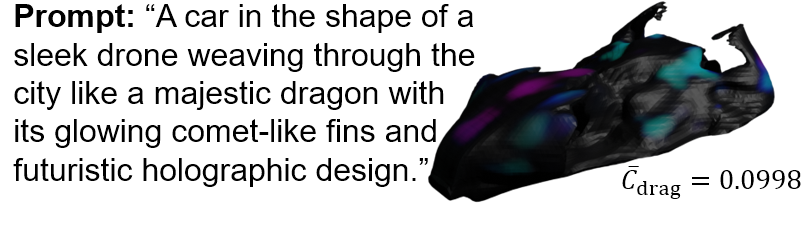} \\
    \includegraphics[width=0.48\linewidth]{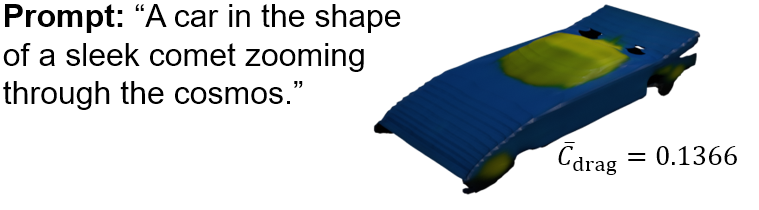} & 
    \includegraphics[width=0.48\linewidth]{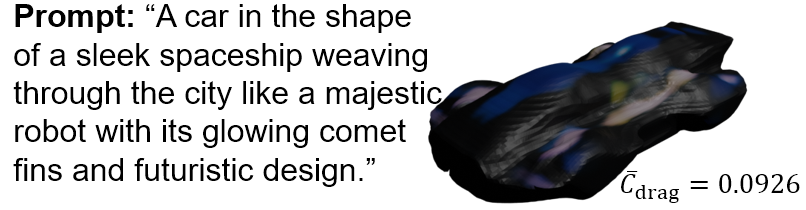} \\
    \includegraphics[width=0.48\linewidth]{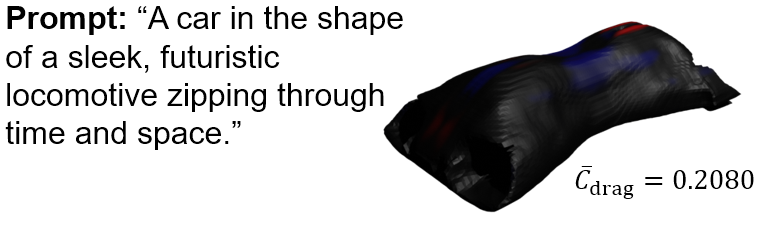} & 
    \includegraphics[width=0.48\linewidth]{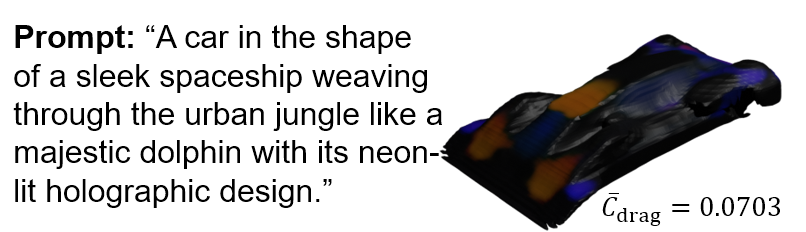} \\
    \includegraphics[width=0.48\linewidth]{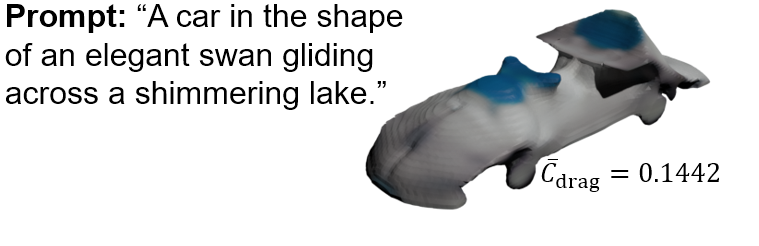} & 
    \includegraphics[width=0.48\linewidth]{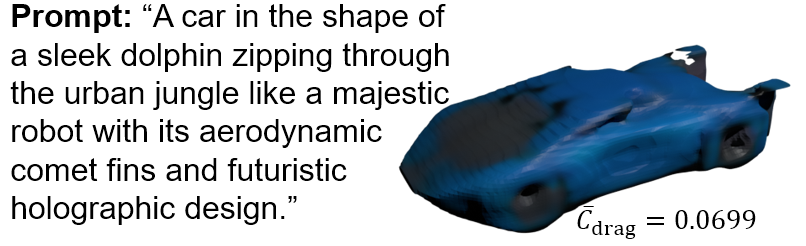} \\
    \includegraphics[width=0.48\linewidth]{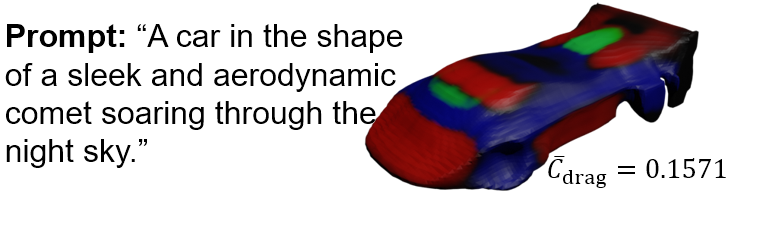} & 
    \includegraphics[width=0.48\linewidth]{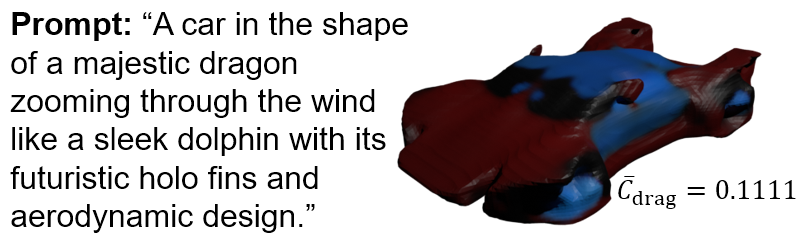} \\
    \includegraphics[width=0.48\linewidth]{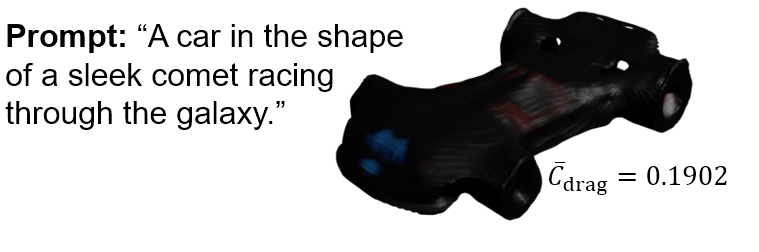} & 
    \includegraphics[width=0.48\linewidth]{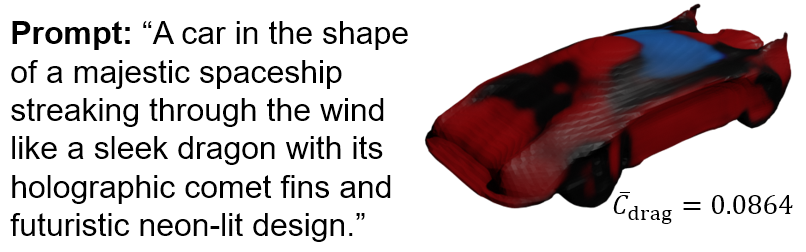} \\
    \textbf{(a) Existing GPT-3.5-Turbo with Shap-E model} & \textbf{(b) LLM-to-Phy3D} \\ [5pt]
  \end{tabular}
  \caption{Examples of 3D cars generated with existing GPT-3.5-Turbo and Shap-E text-to-3D generative model (Left) and with LLM-to-Phy3D (Right). Note that the lower the aerodynamic drag, the better the physical performance of the generated car.}
\end{figure}

\begin{figure}
  \centering
  \begin{tabular}{cc}
    \includegraphics[width=0.50\linewidth]{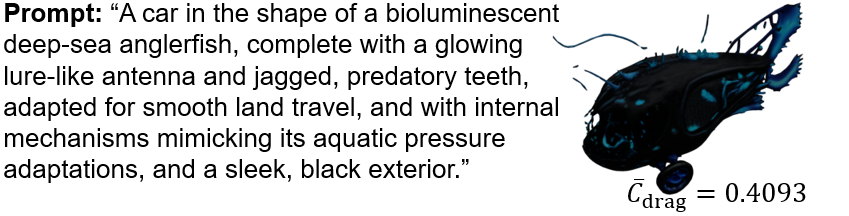} & 
    \includegraphics[width=0.50\linewidth]{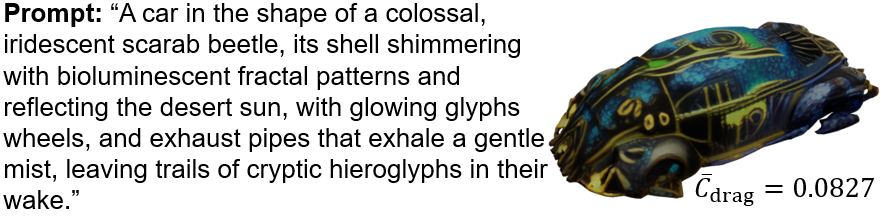} \\
    \includegraphics[width=0.50\linewidth]{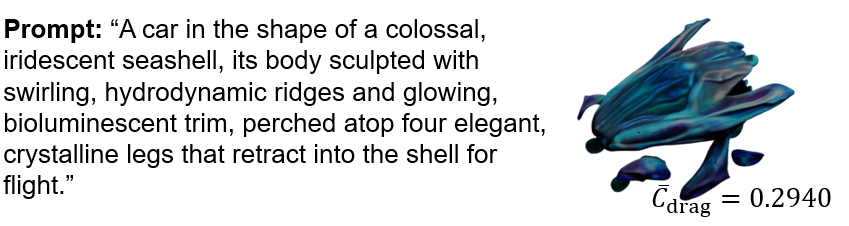} & 
    \includegraphics[width=0.50\linewidth]{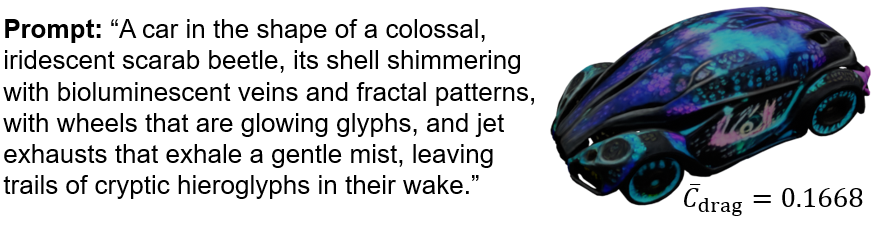} \\
    \includegraphics[width=0.50\linewidth]{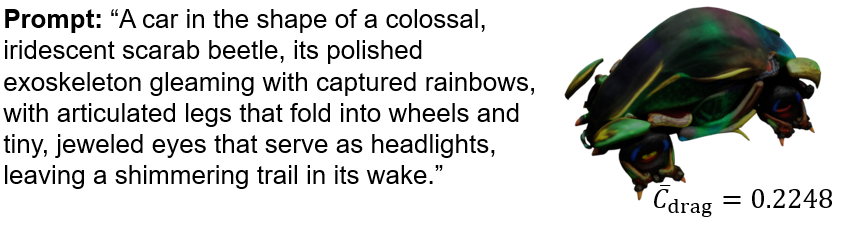} & 
    \includegraphics[width=0.50\linewidth]{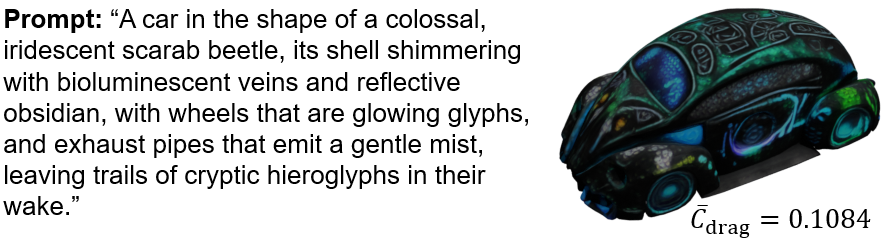} \\
    \includegraphics[width=0.50\linewidth]{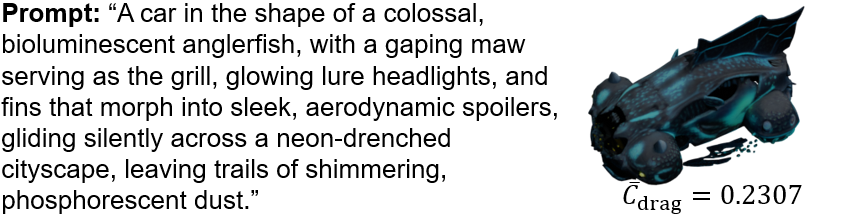} & 
    \includegraphics[width=0.50\linewidth]{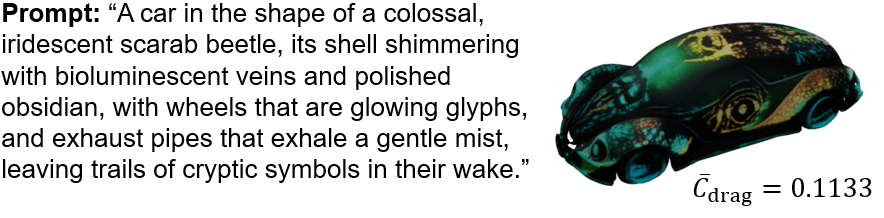} \\
    \includegraphics[width=0.50\linewidth]{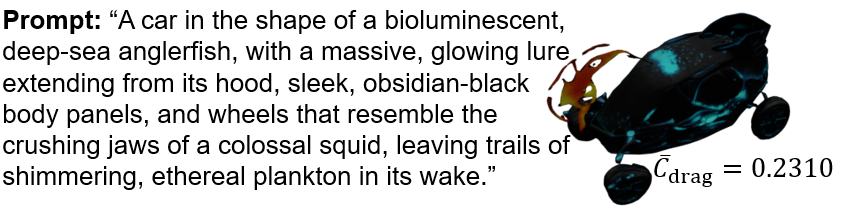} & 
    \includegraphics[width=0.50\linewidth]{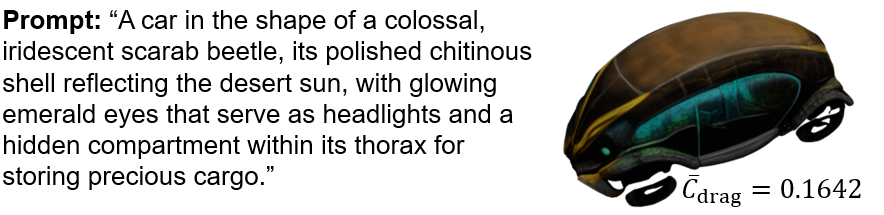} \\
    \includegraphics[width=0.50\linewidth]{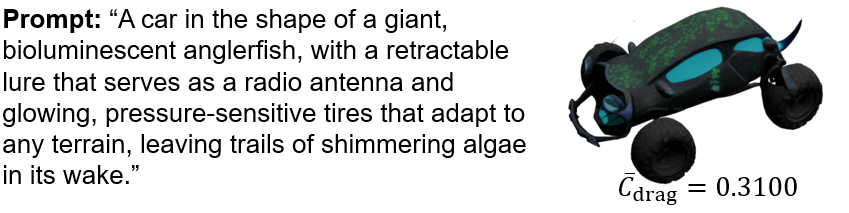} & 
    \includegraphics[width=0.50\linewidth]{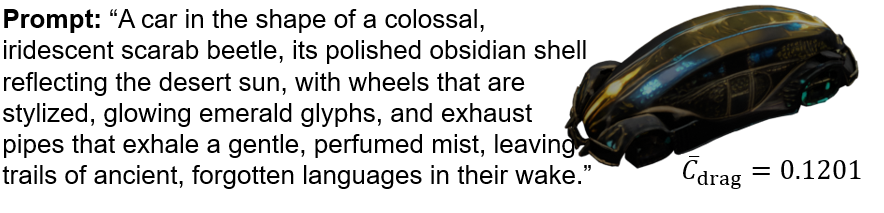} \\
    \includegraphics[width=0.50\linewidth]{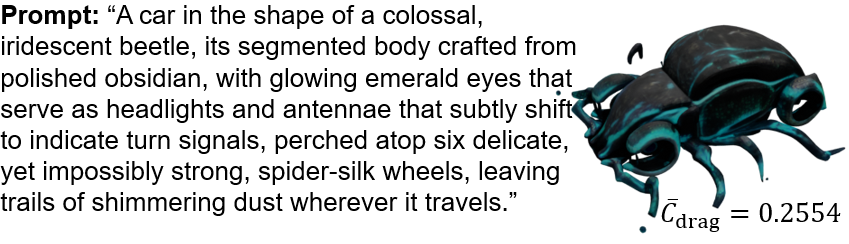} & 
    \includegraphics[width=0.50\linewidth]{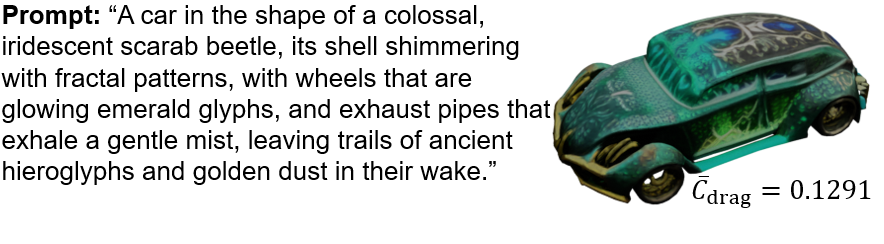} \\
    \includegraphics[width=0.50\linewidth]{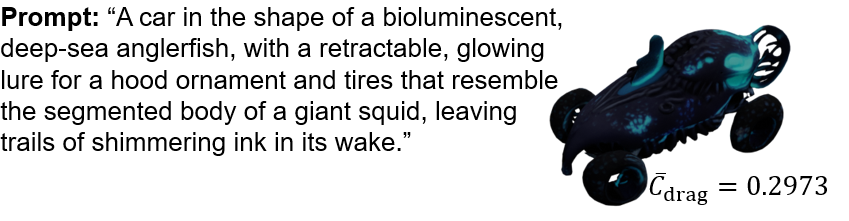} & 
    \includegraphics[width=0.50\linewidth]{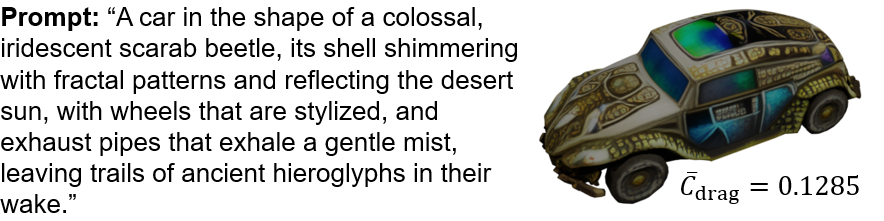} \\
    \includegraphics[width=0.50\linewidth]{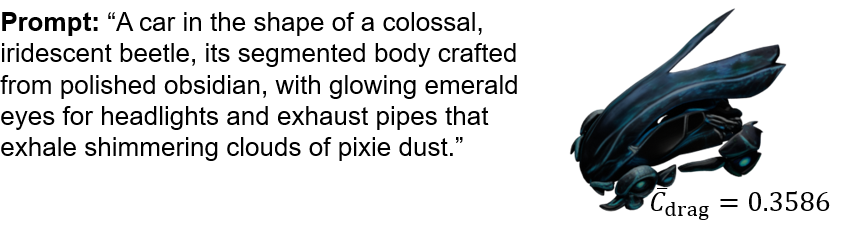} & 
    \includegraphics[width=0.50\linewidth]{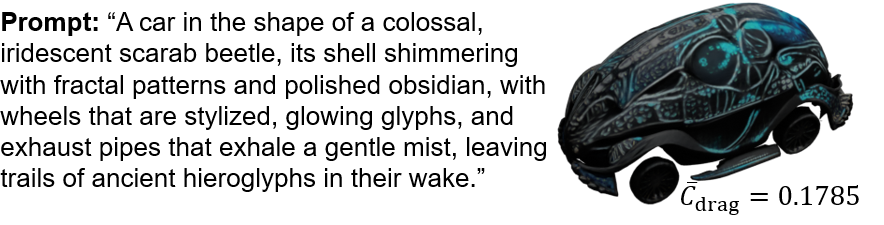} \\
    \textbf{(a) Existing Gemini-2.0-Lite with Trellis model} & \textbf{(b) LLM-to-Phy3D} \\ [5pt]
  \end{tabular}
  \caption{Examples of 3D cars generated with existing Gemini-2.0-Lite and Trellis text-to-3D generative model (Left) and with LLM-to-Phy3D (Right). Note that the lower the aerodynamic drag, the better the physical performance of the generated car.}
\end{figure}

\begin{figure}
  \centering
  \begin{tabular}{cc}
    \includegraphics[width=0.50\linewidth]{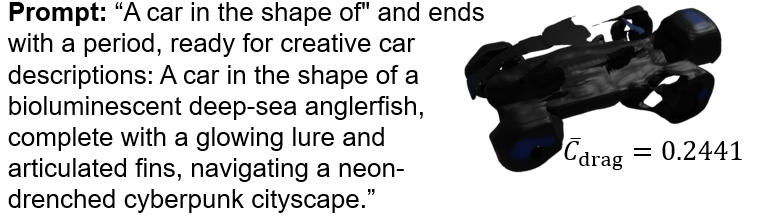} & 
    \includegraphics[width=0.50\linewidth]{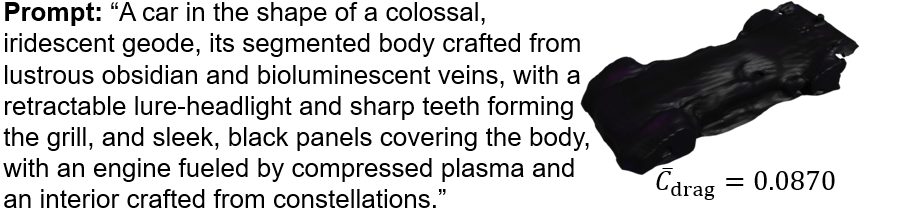} \\
    \includegraphics[width=0.50\linewidth]{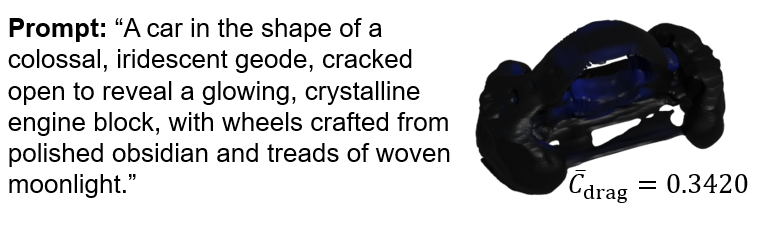} & 
    \includegraphics[width=0.50\linewidth]{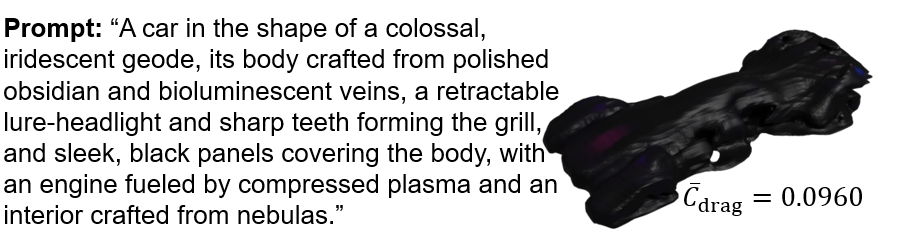} \\
    \includegraphics[width=0.50\linewidth]{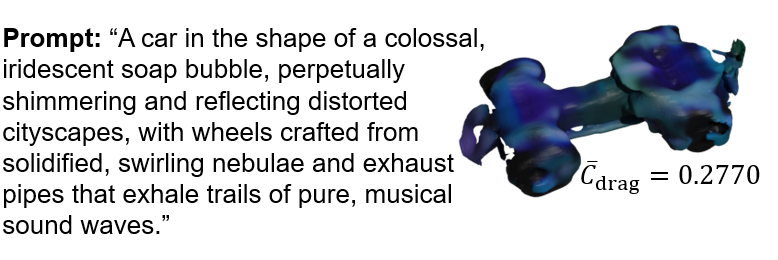} & 
    \includegraphics[width=0.50\linewidth]{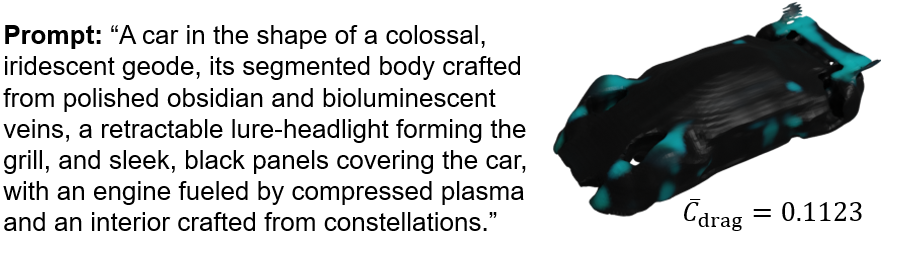} \\
    \includegraphics[width=0.50\linewidth]{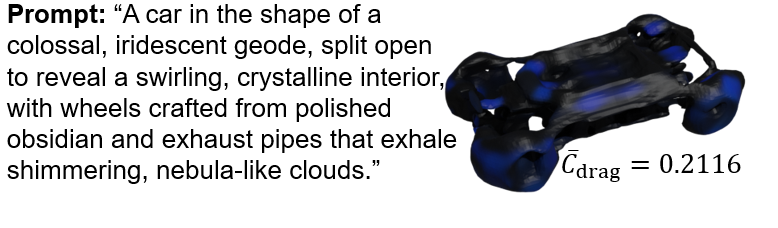} & 
    \includegraphics[width=0.50\linewidth]{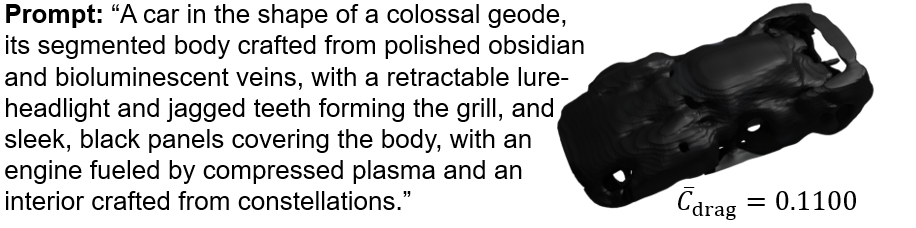} \\
    \includegraphics[width=0.50\linewidth]{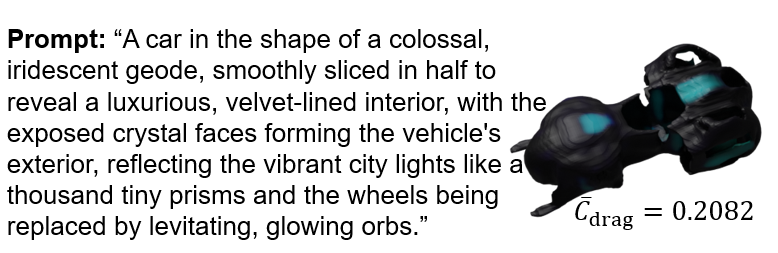} & 
    \includegraphics[width=0.50\linewidth]{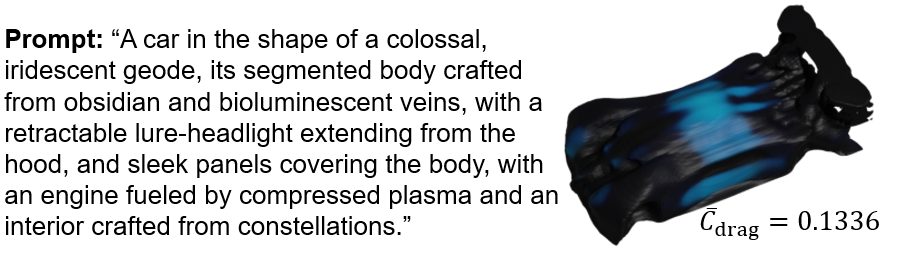} \\
    \includegraphics[width=0.50\linewidth]{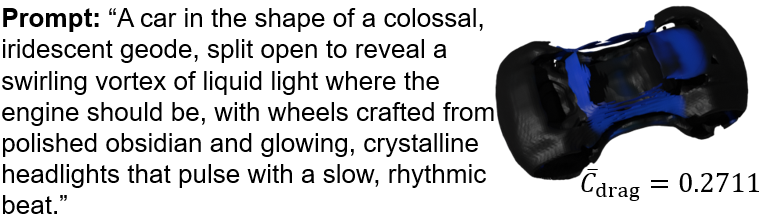} & 
    \includegraphics[width=0.50\linewidth]{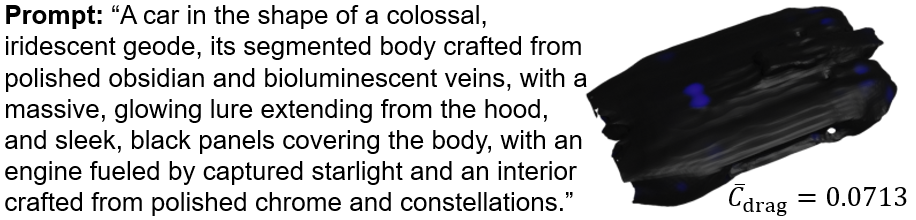} \\
    \includegraphics[width=0.50\linewidth]{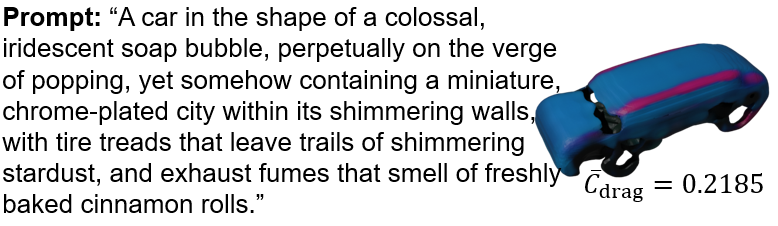} & 
    \includegraphics[width=0.50\linewidth]{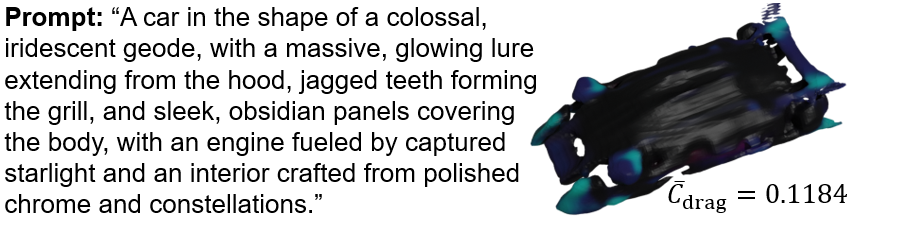} \\
    \includegraphics[width=0.50\linewidth]{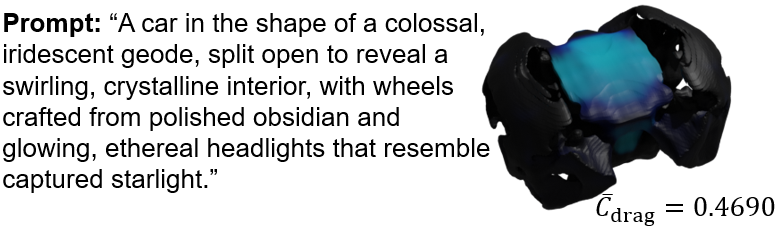} & 
    \includegraphics[width=0.50\linewidth]{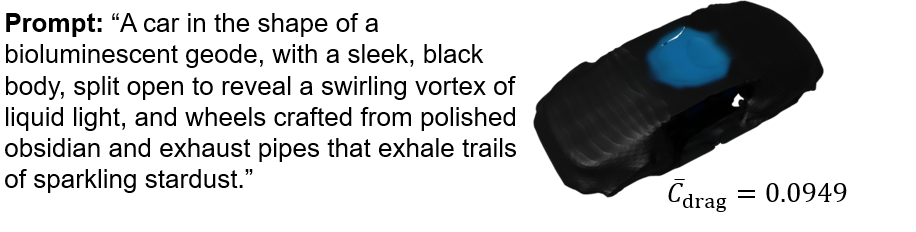} \\
    \includegraphics[width=0.50\linewidth]{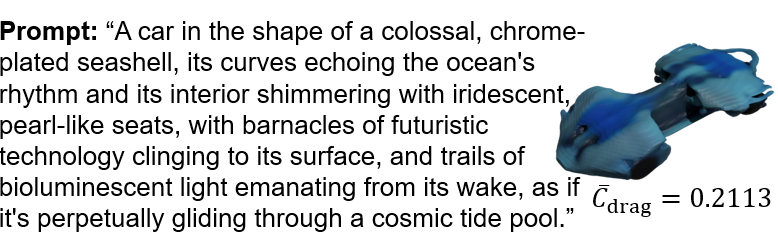} & 
    \includegraphics[width=0.50\linewidth]{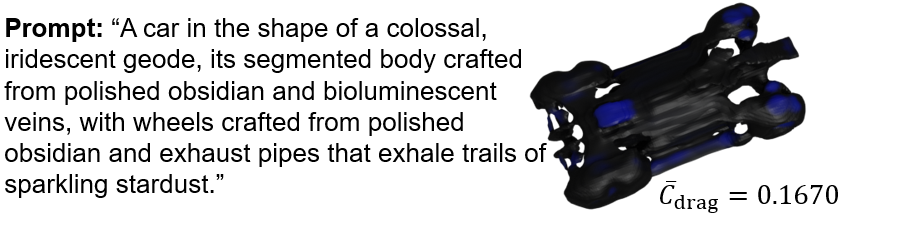} \\
    \textbf{(a) Existing Gemini-2.0-Lite with Shap-E model} & \textbf{(b) LLM-to-Phy3D} \\ [5pt]
  \end{tabular}
  \caption{Examples of 3D cars generated with existing Gemini-2.0-Lite and Shap-E text-to-3D generative model (Left) and with LLM-to-Phy3D (Right). Note that the lower the aerodynamic drag, the better the physical performance of the generated car.}
\end{figure}

\begin{figure}
  \centering
  \begin{tabular}{cc}
    \includegraphics[width=0.50\linewidth]{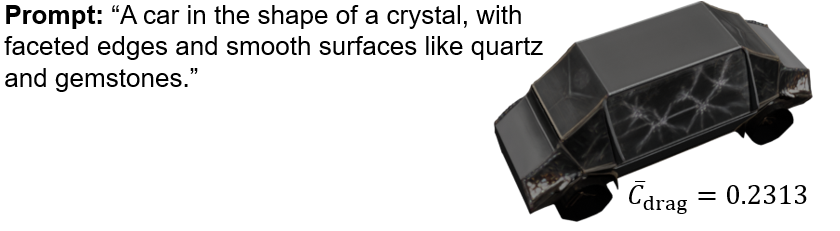} & 
    \includegraphics[width=0.50\linewidth]{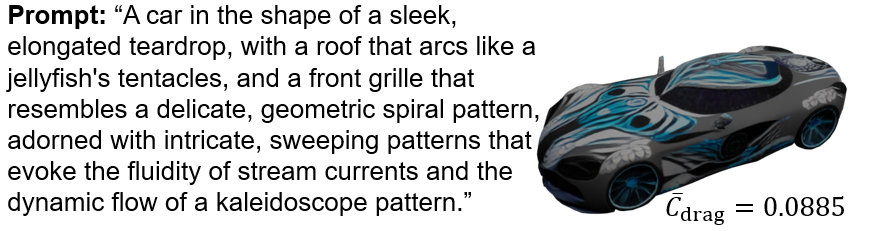} \\
    \includegraphics[width=0.50\linewidth]{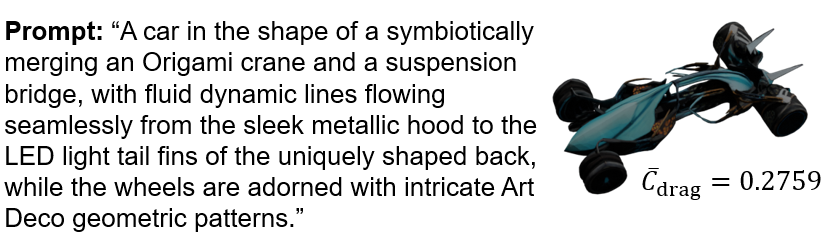} & 
    \includegraphics[width=0.50\linewidth]{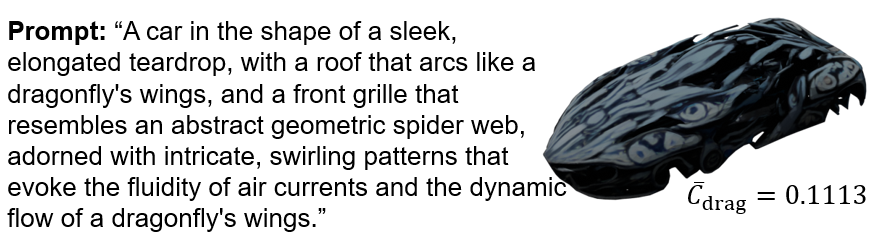} \\
    \includegraphics[width=0.50\linewidth]{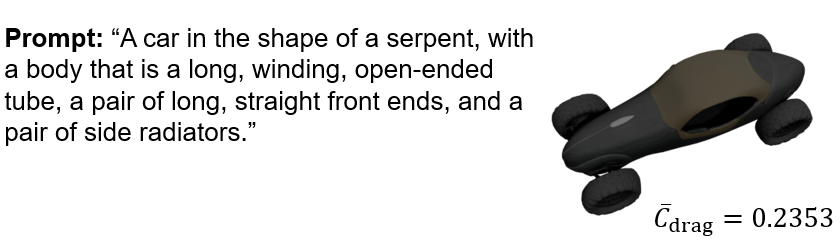} & 
    \includegraphics[width=0.50\linewidth]{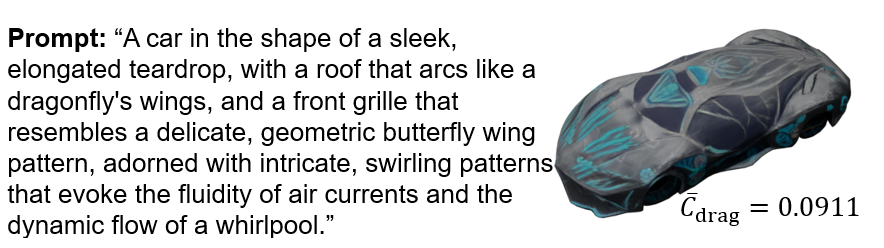} \\
    \includegraphics[width=0.50\linewidth]{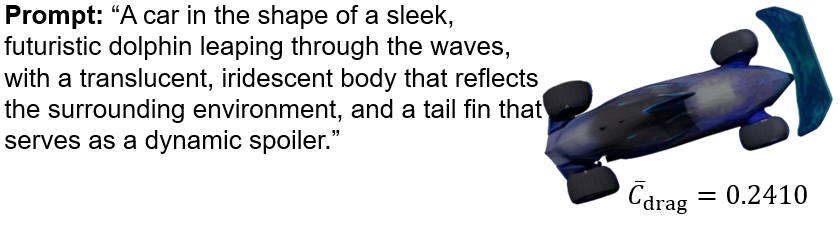} & 
    \includegraphics[width=0.50\linewidth]{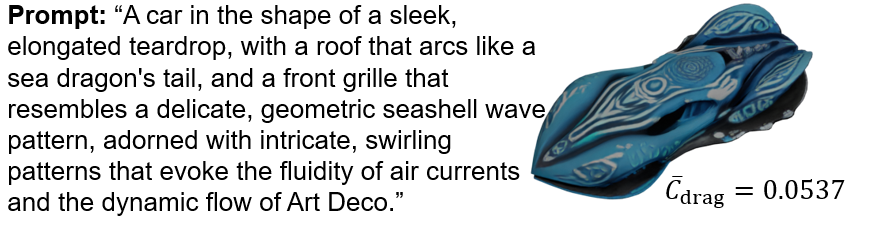} \\
    \includegraphics[width=0.50\linewidth]{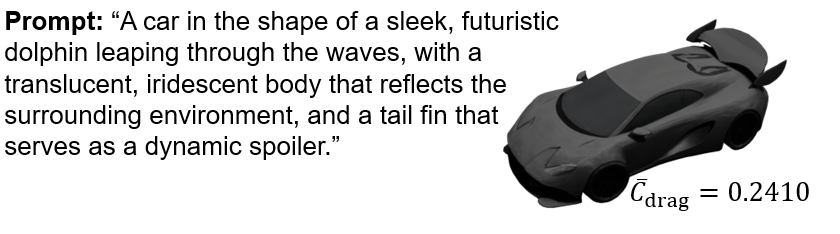} & 
    \includegraphics[width=0.50\linewidth]{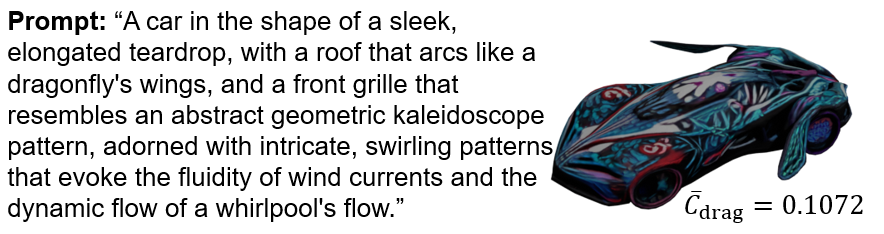} \\
    \includegraphics[width=0.50\linewidth]{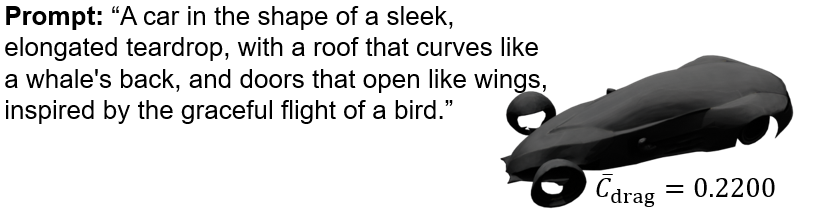} & 
    \includegraphics[width=0.50\linewidth]{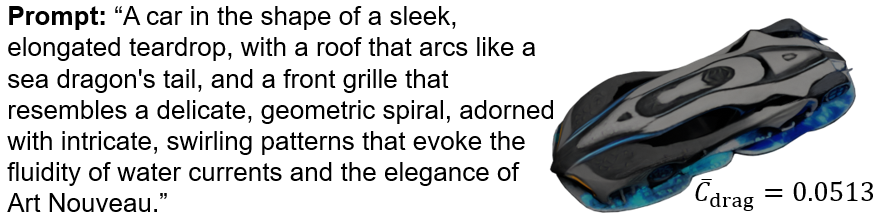} \\
    \includegraphics[width=0.50\linewidth]{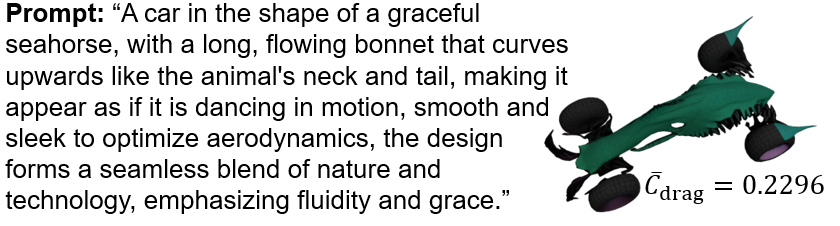} & 
    \includegraphics[width=0.50\linewidth]{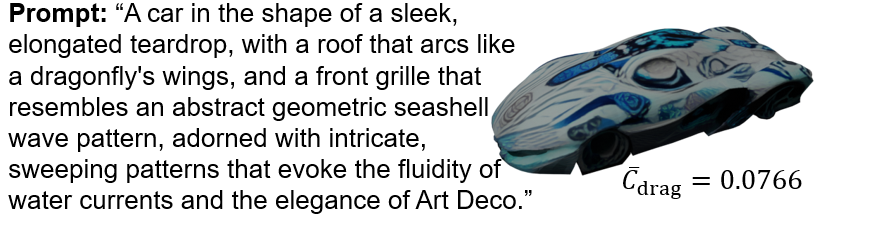} \\
    \includegraphics[width=0.50\linewidth]{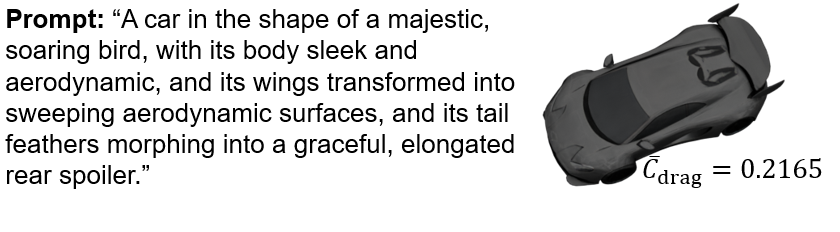} & 
    \includegraphics[width=0.50\linewidth]{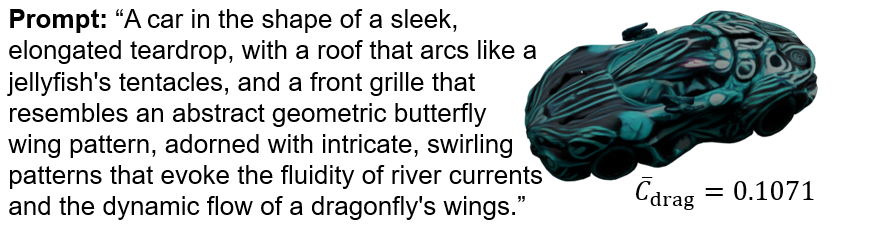} \\
    \includegraphics[width=0.50\linewidth]{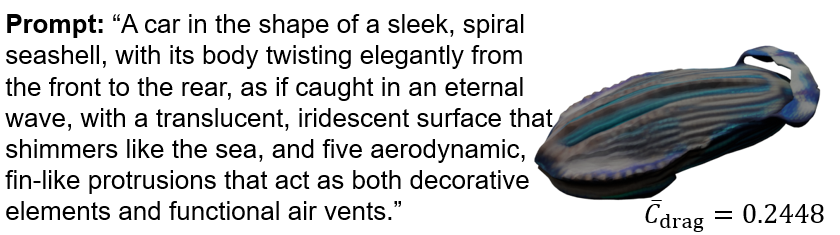} & 
    \includegraphics[width=0.50\linewidth]{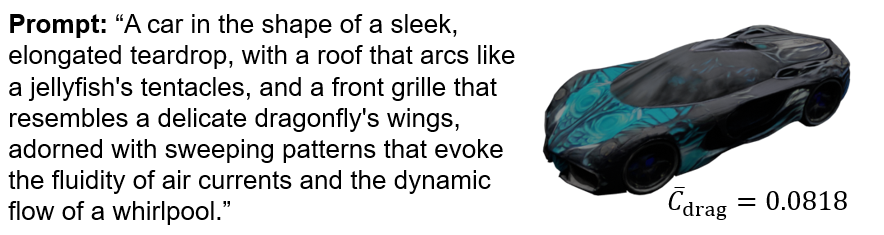} \\
    \textbf{(a) Existing Mistral-3.1-Small with Trellis model} & \textbf{(b) LLM-to-Phy3D} \\ [5pt]
  \end{tabular}
  \caption{Examples of 3D cars generated with existing Mistral-3.1-Small and Trellis text-to-3D generative model (Left) and with LLM-to-Phy3D (Right). Note that the lower the aerodynamic drag, the better the physical performance of the generated car.}
\end{figure}

\begin{figure}
  \centering
  \begin{tabular}{cc}
    \includegraphics[width=0.50\linewidth]{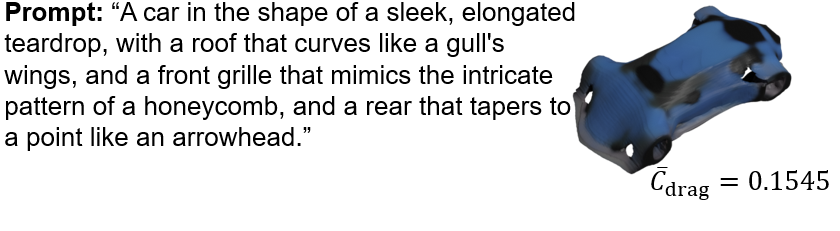} & 
    \includegraphics[width=0.50\linewidth]{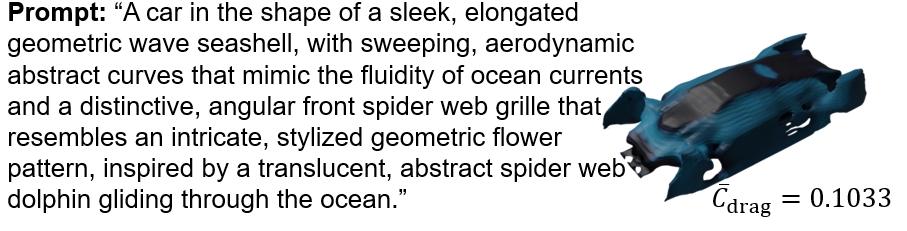} \\
    \includegraphics[width=0.50\linewidth]{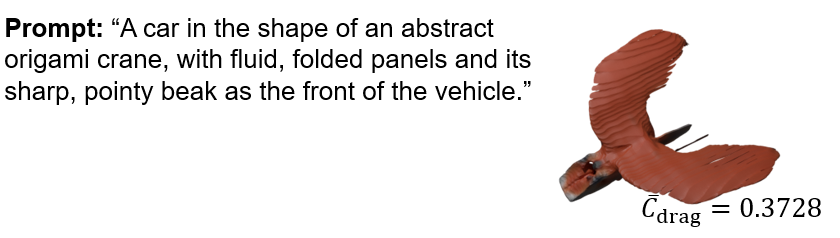} & 
    \includegraphics[width=0.50\linewidth]{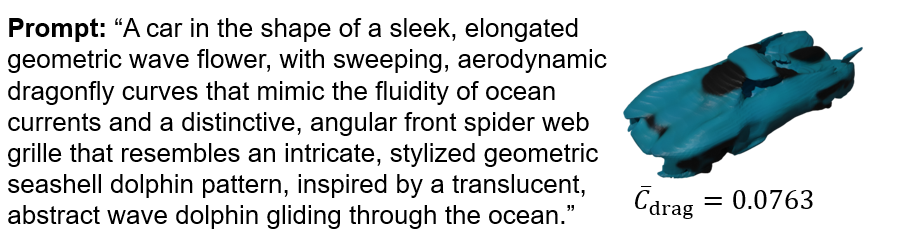} \\
    \includegraphics[width=0.50\linewidth]{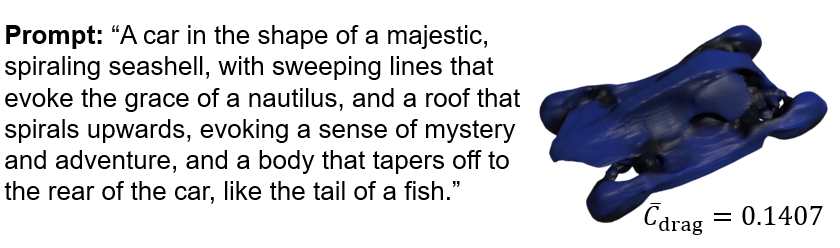} & 
    \includegraphics[width=0.50\linewidth]{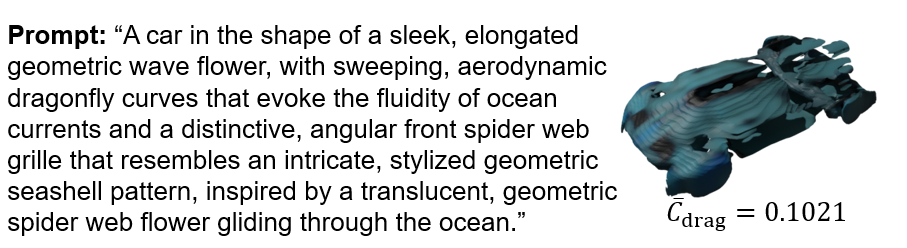} \\
    \includegraphics[width=0.50\linewidth]{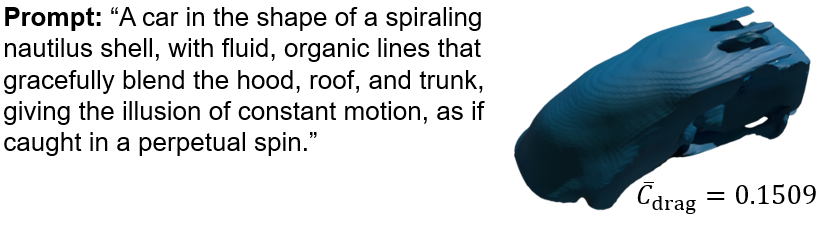} & 
    \includegraphics[width=0.50\linewidth]{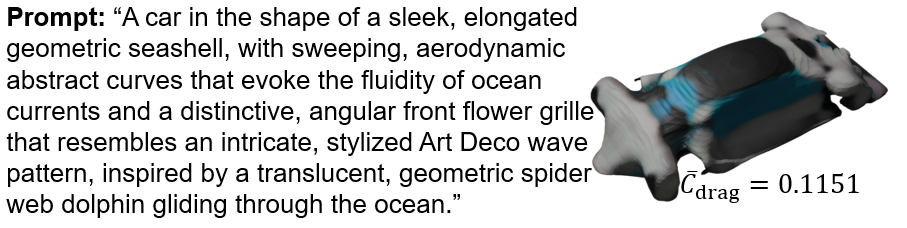} \\
    \includegraphics[width=0.50\linewidth]{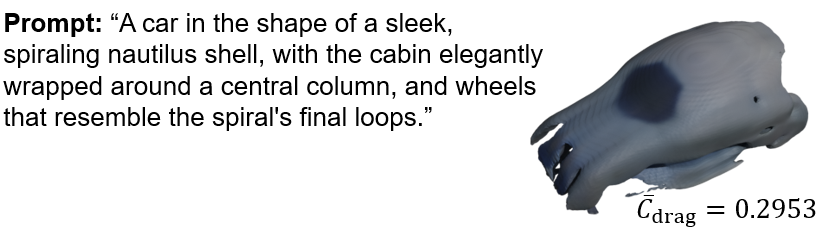} & 
    \includegraphics[width=0.50\linewidth]{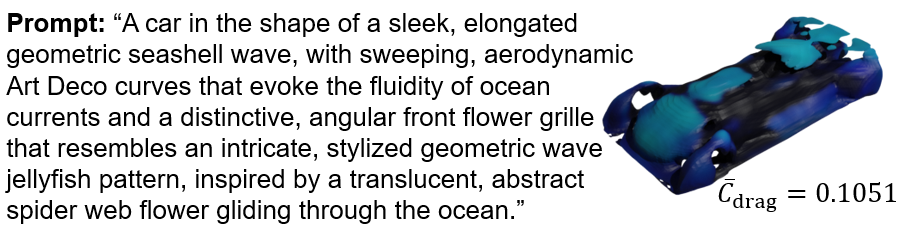} \\
    \includegraphics[width=0.50\linewidth]{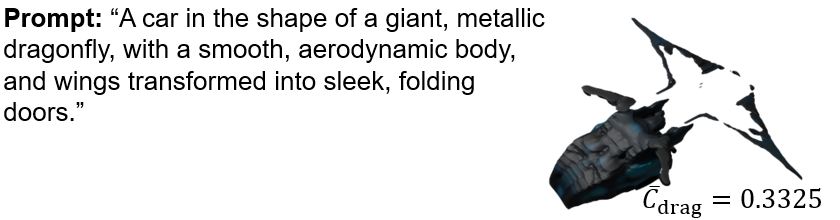} & 
    \includegraphics[width=0.50\linewidth]{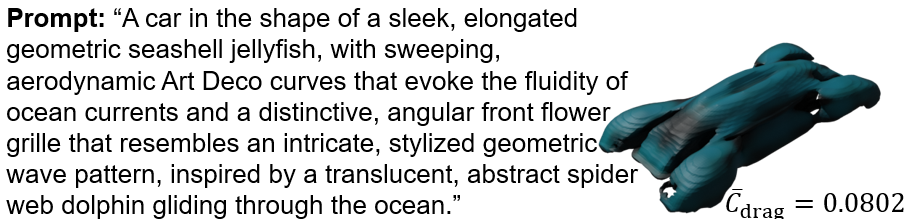} \\
    \includegraphics[width=0.50\linewidth]{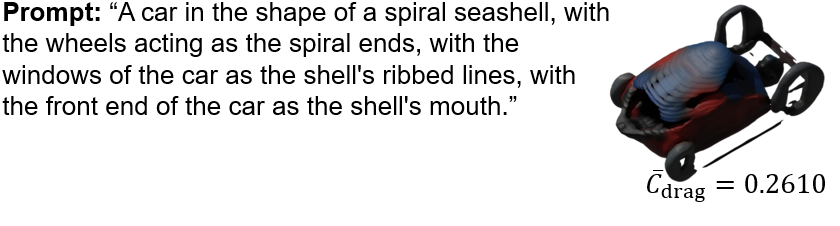} & 
    \includegraphics[width=0.50\linewidth]{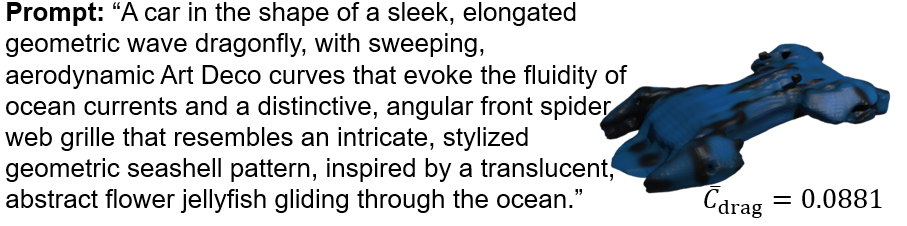} \\
    \includegraphics[width=0.50\linewidth]{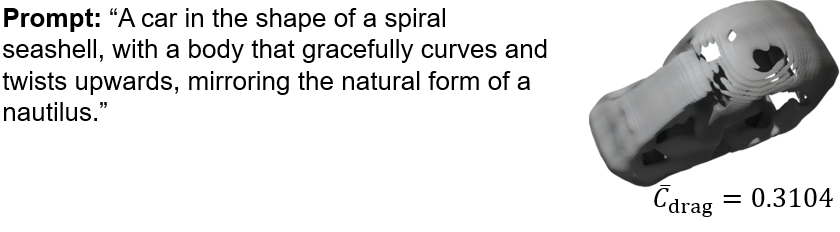} & 
    \includegraphics[width=0.50\linewidth]{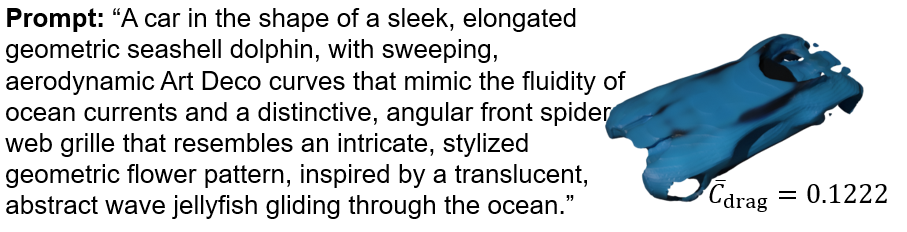} \\
    \includegraphics[width=0.50\linewidth]{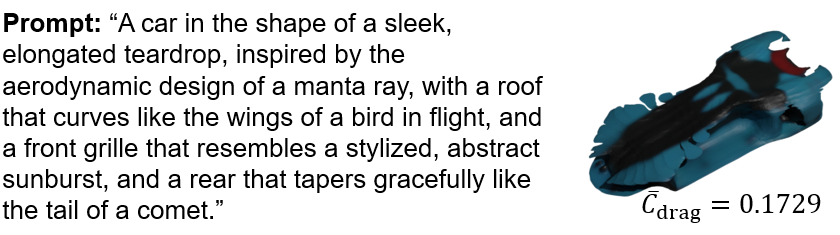} & 
    \includegraphics[width=0.50\linewidth]{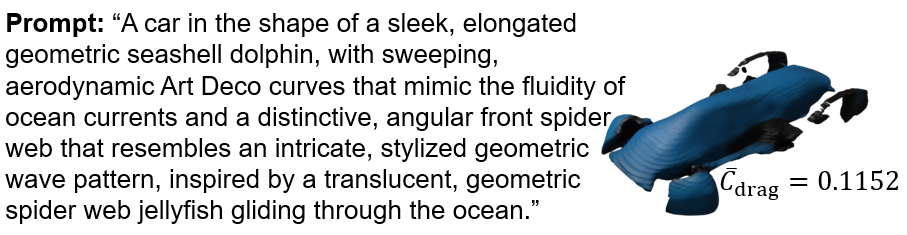} \\
    \textbf{(a) Existing Mistral-3.1-Small with Shap-E model} & \textbf{(b) LLM-to-Phy3D} \\ [5pt]
  \end{tabular}
  \caption{Examples of 3D cars generated with existing Mistral-3.1-Small and Shap-E text-to-3D generative model (Left) and with LLM-to-Phy3D (Right). Note that the lower the aerodynamic drag, the better the physical performance of the generated car.}
\end{figure}





\end{document}